%% file: colm2026_conference.tex
\documentclass{article} 
\usepackage[final]{colm2026_conference}
\usepackage{float}
\usepackage[most]{tcolorbox}
\usepackage{microtype}
\usepackage{amsmath}
\usepackage{amssymb}
\usepackage{graphicx}
\usepackage{hyperref}
\usepackage{url}
\usepackage{booktabs}
\usepackage{wrapfig}
\usepackage{enumitem}
\usepackage[ruled,vlined,linesnumbered]{algorithm2e}
\SetAlgoNlRelativeSize{0}

\usepackage{lineno}

\definecolor{darkblue}{rgb}{0, 0, 0.5}
\hypersetup{colorlinks=true, citecolor=darkblue, linkcolor=darkblue, urlcolor=darkblue}

\title{Learning as Reasoning Unfolds: Progressive Rollout Allocation for Efficient Reinforcement Learning}

\author{Heyang Jiang\thanks{Correspondence to: \texttt{jianghy0581@gmail.com}}  ,   Henry Liu, Baharan Mirzasoleiman \\
University of California, Los Angeles
}

\begin{document}

\ifcolmsubmission
\linenumbers
\fi

\maketitle

\begin{abstract}
Reinforcement learning with verifiable rewards (RLVR) has emerged as a highly effective framework for improving LLM reasoning, with methods such as GRPO among its most successful instantiations. However, GRPO relies on repeated generation of long chain-of-thought rollouts. Training time scales with the number of rollouts, a large fraction of which are uninformative. Thus,
GRPO is computationally expensive and unstable.
To mitigate this, existing approaches either generate a larger pool of rollouts and filter the most informative prompts, or leverage historical signals for filtering at later stages of training. These strategies offer modest performance gains, but slow down the overall process.
To address this, we propose VarIance Guided Online Rollout allocation (VIGOR) which instead of allocating a fixed rollout budget per example,  begins with a small number of rollouts for all examples in a batch and iteratively allocates additional rollouts to those with the highest group reward variance until a fixed total rollout budget is reached.
Theoretically, we show that under RLVR, reward variance controls the gradient magnitude, and derive VIGOR's closed-form speedup ratio over GRPO, which grows with refinement rounds under Pareto-distributed reward variance. Experiments on mathematical reasoning and coding tasks show that VIGOR reaches target accuracy with up to 2.3$\times$ fewer rollouts on math, reaches GRPO's final coding full pass rate with 1.49$\times$ fewer rollouts, and improves the coding average test pass rate by 3.4 points.
\end{abstract}

\begin{figure}[h]
\centering
\includegraphics[width=0.9\linewidth]{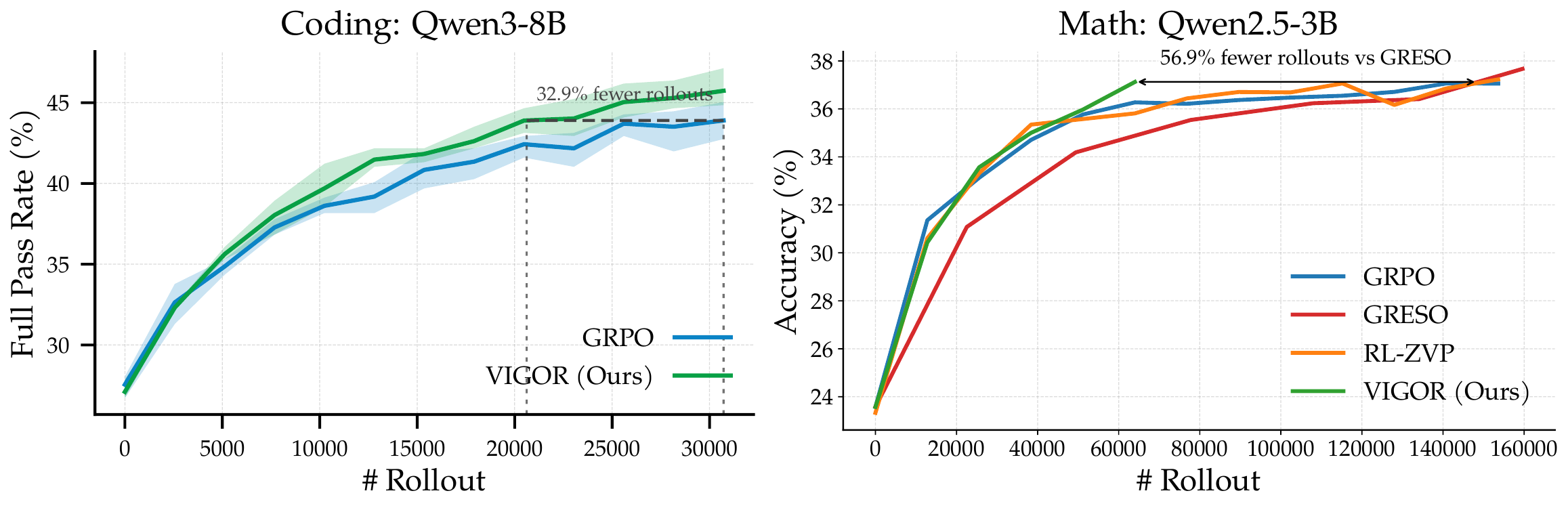}
\caption{Left: VIGOR Coding results substantially outperform GRPO baseline on Live CodeBench v6 with Qwen3-8B, achieving faster full pass rate improvement across rollout budgets, requiring 1.49$\times$ fewer rollouts to reach GRPO's final full pass rate. Shaded regions show the standard deviation across 3 seeds. Right: VIGOR Math results comparing against GRPO, GRESO, and RL-ZVP baselines when training Qwen2.5-3B on MATH. y-axis shows the average score on six reasoning benchmarks, including Math500, AIME24, AMC, Minerva Math, Gaokao, and Olympiad Bench. VIGOR consistently achieves faster improvement under the same rollout budget, requiring 2.3$\times$ fewer rollouts to reach target accuracy over all the other baselines.}
\label{fig:score_vs_rollouts}
\end{figure}

\input{sec/01_intro}
\input{sec/02_related}
\input{sec/03_preliminaries}
\input{sec/04_observations}

\input{sec/05_method}
\input{sec/06_experiments}
\input{sec/07_conclusion}

\clearpage


\section*{Acknowledgments}
This research was supported by the NSF-Simons AI Institute for Cosmic Origins (CosmicAI), NSF CAREER Award 2146492, and NSF AI Institute for Foundations of Machine Learning (IFML).


\bibliography{colm2026_conference}
\bibliographystyle{colm2026_conference}

\clearpage
\appendix
\input{sec/appendix}

\end{document}

%% file: sec/01_intro.tex
\section{Introduction}
\label{sec:intro}

Recent large language models \citep{openai2024o1systemcard,guo2025deepseekr1,yang2024qwen25,yang2024qwen25math} have shown that strong reasoning abilities can be effectively elicited through reinforcement learning with verifiable rewards (RLVR), which enables learning from automatically verifiable signals over chain-of-thought reasoning trajectories rather than relying on curated supervision. Among these approaches, Group Relative Policy Optimization (GRPO) \citep{shao2024deepseekmath} has emerged as a particularly efficient variant, improving the policy by sampling multiple rollouts per prompt and updating it based on their relative rewards within each group. However, GRPO relies on generating a large number of rollouts, many of which are uninformative due to having little or no reward variance across samples. This not only slows down training \citep{lambert2025tulu3}---e.g. 7B-scale runs can take multiple days---but can also lead to instability in optimization. Consequently, a central challenge is to identify and prioritize the most informative examples for GRPO \citep{li2025limr,zheng2025greso,mao2026dps}, with the goal of improving both training efficiency and overall performance.

Conventional data selection methods, 
such as gradient- or loss-based methods designed for supervised learning \citep{nguyen2024minibatchcoresets,zhou2025davir,chen2024alpagasus} are not applicable to RLVR due to its very different learning objective. 
To enhance GRPO, recent works have mainly explored two directions. 
The first direction, initiated by DAPO~\citep{xiao2025dapo} and extended by later works~\citep{zheng2025greso,mao2026dps}, 
either generate a larger number of rollouts for every prompt and dynamically sample difficult prompts with non-zero group reward variance, or use rollout statistics in the early training stage to sample difficult prompts afterwards.
However, the first approach incurs a large computational overhead and the latter yields marginal improvement.
The second direction improves rollout utilization efficiency through auxiliary signals or stale rollouts~\citep{zvp2026,xu2025pods,zheng2025m2po}. While these methods can increase data efficiency, they modify the training objective and 
have limited efficiency gains relative to their added complexity.\looseness=-1

In this work, we propose an effective method to enhance the efficiency of GRPO. First, we make two key observations: (1) rollout generation is a primary driver of training cost when training reasoning models, and (2) the early stages of training contribute disproportionately to the final performance gains. These observations highlight the need for a data selection strategy that allocates rollout budget adaptively—focusing computation \textit{only} on the most informative prompts without requiring expensive generation and subsequent filtering of large numbers of rollouts. Such adaptivity is especially critical in the early stages of training, where efficient use of compute has the greatest impact on overall performance.


Motivated by the above insights, we propose VIGOR, an iterative data selection framework for GRPO that focuses compute on the most informative examples (Figure~\ref{fig:overview}). 
We begin by showing that, under a binary reward setting, the magnitude of the GRPO gradient is directly governed by the within-group reward variance, revealing a natural signal for identifying informative prompts.
Building on this observation, VIGOR departs from the standard practice of allocating a large rollout budget uniformly across all training instances. Instead, it starts with a small number of rollouts per prompt and iteratively identifies examples with the highest group-level reward variance, selectively increasing their rollouts until a fixed budget is reached. Unlike prior approaches that rely on static or historical heuristics to filter low-value samples, VIGOR adaptively exploits signals that emerge during training, without modifying the underlying GRPO paradigm. This design avoids the additional rollout overhead incurred by methods that generate extra samples solely for filtering purposes. Moreover, we show that this variance-guided allocation yields a closed-form speedup ratio over GRPO that grows exponentially with the number of refinement rounds under a Pareto-distributed reward-variance model, substantially enhancing training efficiency and performance.


We evaluate VIGOR on six mathematical reasoning benchmarks across Qwen2.5-1.5B, Qwen2.5-3B, Qwen2.5-7B~\citep{yang2024qwen25}, and Phi-4-Mini-Instruct~\citep{phi4mini2025}, as well as on LiveCodeBench v6 with Qwen3-8B~\citep{yang2025qwen3}. On math, VIGOR consistently reaches the same accuracy with substantially fewer rollouts, with up to a 2.3$\times$ rollout-efficiency improvement on Qwen2.5-3B, and achieves the best average score under the fair rollout-matched comparison across all four model settings. On coding, VIGOR reaches GRPO's final full pass rate with 1.49$\times$ fewer rollouts and improves the average test pass rate by 3.4 points. These results show that VIGOR is a simple and effective rollout allocation strategy for improving RLVR efficiency across both math and coding tasks.


%% file: sec/02_related.tex
\section{Related Work}
\label{sec:related}

\paragraph{Reinforcement learning for LLM reasoning.}
Reinforcement learning has long become a powerful tool for enabling LLMs to generate more contextually appropriate and ethically-grounded responses by incorporating fine-grained preference signals into the optimization objective, aligning their responses with human-centric quality metrics ~\citep{christiano2017deep,stiennon2020learning,ouyang2022training}.  
Build upon this foundation, Reinforcement Learning from Verifiable Rewards (RLVR) has been proposed to replace potentially biased human feedback with objective, deterministic reward signals. By leveraging verifiable outcomes, RLVR significantly enhances reasoning capabilities in high-precision tasks such as mathematics and code generation ~\citep{lightman2023lets,shao2024deepseekmath,guo2025deepseekr1}.
This reasoning advancement relies on the evolution of policy optimization methods from Proximal Policy Optimization (PPO)~\citep{schulman2017proximal} to Group Relative Policy Optimization (GRPO)~\citep{shao2024deepseekmath}. By replacing the value function with group-relative rewards, GRPO achieves superior training efficiency and establishes a new paradigm for scalable alignment.
Since then, GRPO has emerged as a standard paradigm for industrial-scale post-training of reasoning LLMs and a cornerstone of academic research on RLVR~\citep{guo2025deepseekr1,xiao2025dapo,liu2025drgrpo,deepseekv3.2}.

\paragraph{Data selection and efficiency for RLVR}
Data-centric strategies have emerged as an important direction for improving the stability and efficiency of RLVR training.
For example, LIMR~\citep{li2025limr} 
finds examples that their learning progress correlates best with the model’s overall improvement during training on full data,
enabling a 7B model to substantially outperform SFT-oriented baselines such as s1~\citep{muennighoff2025s1} and LIMO~\citep{ye2025limo}.
The importance of data quality is further highlighted in GRPO-based training. For instance, DAPO~\citep{xiao2025dapo} generates a larger number of rollouts for every example and samples difficult examples with non-zero relative advantage.
However, generating a larger number of rollouts to find the difficult examples is expensive and slows down trainig.
To improve training throughput, GRESO~\citep{zheng2025greso} and DPS~\citep{mao2026dps} leverage statistics of the group rewards in early stage of training to sample difficult examples in later training stages. While being more efficient than DAPO, these approaches yield marginal performance improvement as they cannot focus training on difficult examples during the initial phase of training which contributes the most to performance.
In parallel, complementary approaches 
such as RL-ZVP~\citep{zvp2026} repurposes zero-advantage samples with entropy-based signals.  PODS~\citep{xu2025pods} generate a larger number of rollouts and selects a subset that provide the most diverse reward signals. However, this incurs very high computational costs for reasoning with long chain of thoughts.
M2PO~\citep{zheng2025m2po} leverages stale rollouts to reduce the rollout generation costs.

Recent adaptive-rollout methods further address the limitations of prior filtering-based methods by dynamically allocating rollout budgets during RLVR. VIP~\citep{nguyen2026vip} estimates prompt difficulty from embedding-based success prediction, which enables early allocation but can be unstable when the estimate is noisy. AR3PO~\citep{zhang2025ar3po} and XRPO~\citep{bamba2025xrpo} use iterative rollout allocation within each step, but still focus on improving exploration for difficult prompts.

Compared with prior data-selection and adaptive-rollout methods, VIGOR uses direct online reward variance without extra guidance or additional computation, provides a convergence-acceleration guarantee, and naturally shifts computation toward harder prompts over training. In our experiments, we compare our method, VIGOR, with GRESO which performs difficulty-aware sampling and RL-ZVP which improves rollout utilization efficiency.

%% file: sec/03_preliminaries.tex
\section{Group Relative Policy Optimization (GRPO)}
\label{sec:observations}

In this section, we first review Group Relative Policy Optimization (GRPO)~\citep{shao2024deepseekmath}, which has become a dominant paradigm for RLVR. Then, we present two key observations that motivate our method: (1) rollout generation is a major contributor to training cost across scales, and (2) early stages of training contribute disproportionately to the model's  accuracy gains.

\subsection{Preliminaries}
\label{sec:preliminaries}


GRPO is a policy optimization algorithm for RLVR that replaces value-function estimation with group-relative rewards. For each prompt $x \sim \mathcal{D}$, the current policy $\pi_\theta$ samples a group of $G$ responses $\{y_1, \dots, y_G\}$, each receiving a reward $r_i = R(x, y_i)$. The relative advantage is computed within the group as
\begin{equation}
    A_i = \frac{r_i - \bar{r}}{\sigma_r + \delta},
    \quad
    \bar{r} = \frac{1}{G} \sum_{j=1}^{G} r_j,
    \label{eq:grpo_advantage}
\end{equation}
where $\bar{r}$ and $\sigma_r$ are the group-wise mean and standard deviation of rewards, and $\delta$ is a small numerical stabilizer. The policy is then updated with a PPO-style clipped objective:
\begin{equation}
    \mathcal{L}_{\mathrm{GRPO}}(\theta)
    = \mathbb{E}_{x, \{y_i\}}
    \left[
    \frac{1}{G} \sum_{i=1}^{G}
    \min\left(
        \rho_i(\theta) A_i,
        \mathrm{clip}(\rho_i(\theta), 1-\varepsilon, 1+\varepsilon) A_i
    \right)
    \right]
    - \beta \, D_{\mathrm{KL}}\!\left(\pi_\theta \,\|\, \pi_{\mathrm{ref}}\right),
    \label{eq:grpo_objective}
\end{equation}
where $\rho_i(\theta) = \frac{\pi_\theta(y_i \mid x)}{\pi_{\theta_{\mathrm{old}}}(y_i \mid x)}$ is the importance ratio, $\varepsilon$ is the clipping threshold, and $\pi_{\mathrm{ref}}$ is a reference policy.



%% file: sec/04_observations.tex
\subsection{Empirical Observations}
Next, we make two important observations about GRPO.

\subsubsection*{(1) Rollout Generation is a Major Factor in Scaling Training Cost}

\begin{figure*}[t]
\centering
\includegraphics[width=0.8\textwidth]{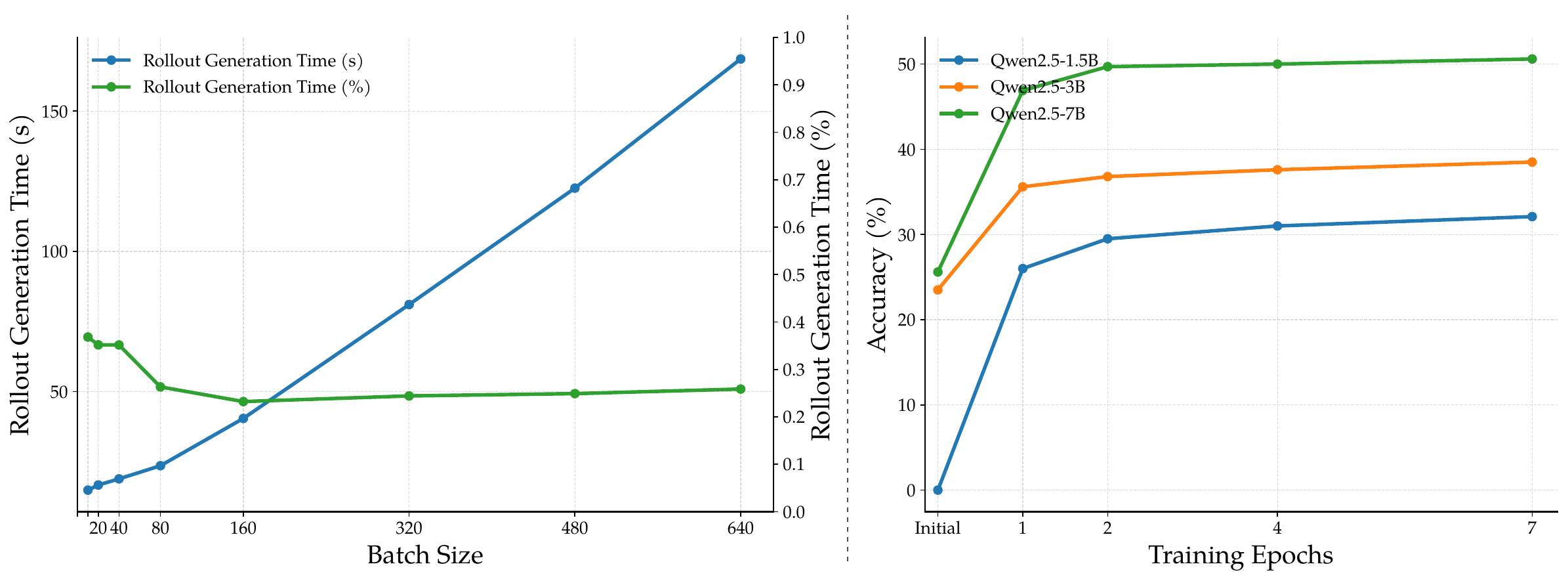}
\caption{Empirical observations motivating VIGOR. \textbf{Left}: rollout generation time as a function of batch size for baseline GRPO with Qwen2.5-1.5B training on 4$\times$A40 GPUs. Once the batch size exceeds a certain threshold, rollout generation time scales linearly with batch size and takes up at least 50\% of total training time, showing that rollout cost is a major and scaling component of RLVR training. \textbf{Right}: empirically, the early stage of training contributes a disproportionate share of the final validation-score improvement, indicating that improving efficiency early in training is vitally important.}
\label{fig:timing_and_val_score}
\end{figure*}

As shown in Figure~\ref{fig:timing_and_val_score} (left), we examine the runtime composition of baseline GRPO training for Qwen2.5-1.5B under RLVR and observe that rollout generation occupies at least 50\% of total training time for various batch sizes, a substantial fraction of the end-to-end training cost, making it a bottleneck that cannot be ignored. This is especially pronounced in long-CoT settings. Moreover, we find that the time required for rollout generation is roughly proportional to the number of rollouts produced. This linear scaling, observed once the batch size exceeds a certain threshold, suggests that the rollout budget can be reallocated and generated more finely in an iterative manner without requiring extra resources.

\subsubsection*{(2) Early-Stage Training Contributes Disproportionately to Performance Gains}

As shown in Figure~\ref{fig:timing_and_val_score} (right), we observe that the early stage of RLVR training contributes a disproportionate share of the final performance improvement. When training on the MATH Dataset~\citep{hendrycks2021math} with 1.5B and 3B models, the first epoch alone can already account for around $80\%$ of the final validation-score gain. Even for the 7B model trained on the larger mixed DAPO~\citep{xiao2025dapo} and MATH~\citep{hendrycks2021math} data, the improvement obtained in the first epoch is comparable to the total gain from the following six epochs. These results suggest that training efficiency at the beginning of RLVR is especially important: if informative training data is not utilized early in training, a large portion of achievable optimization progress may be lost. In fact, many existing data selection 
methods~\citep{zheng2025greso,mao2026dps} rely on difficulty perception from training history and thus cannot filter informative examples until at least the second epoch.



%% file: sec/05_method.tex

\section{Method}
\label{sec:method}

\begin{figure*}[t]
\centering
\includegraphics[width=\textwidth]{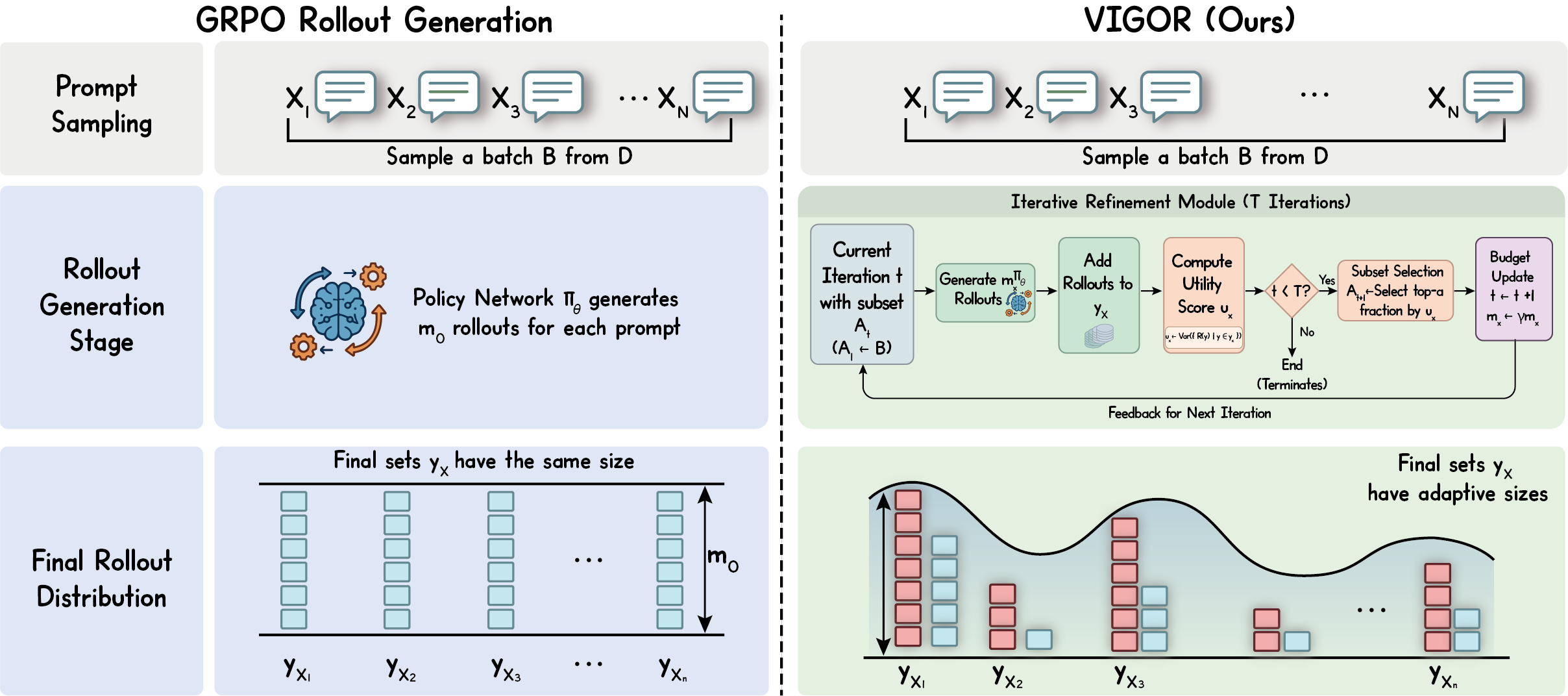}
\caption{Overview of VIGOR. The method performs iterative rollout generation and variance-based prompt selection, progressively reallocating rollout budget toward more informative prompts while keeping the underlying GRPO update pipeline unchanged.}
\label{fig:overview}
\end{figure*}



Next, we propose an iterative rollout generation and maximum-variance selection framework for RLVR. First, in Section~\ref{sec:max_variance_selection}, we prove that the rollout-reward variance of each prompt controls its gradient magnitude and thus identifies informative prompts for GRPO training. Motivated by this, in Section~\ref{sec:iterative_rollout_generation} we introduce VIGOR. Finally, in Section~\ref{sec:convergence_time}, we show that this variance-guided allocation also yields a convergence-timescale reduction under a heavy-tailed prompt-variance model.
Unlike standard GRPO, which generates a fixed number of rollouts per prompt in a single pass, VIGOR begins with a small rollout budget for each prompt in a batch. At each iteration, it retains a progressively smaller subset of prompts with the highest reward variance, allocates additional rollouts to these selected prompts, and merges the new samples into the existing rollout pool before computing advantages over the aggregated set. Aside from rollout generation, the rest of the GRPO training pipeline remains unchanged. By redistributing the original rollout budget across iterations, VIGOR improves training efficiency without increasing the total number of rollouts, while enabling effective identification of the most informative prompts from the earliest stages of training, aligned with our observations in Section~\ref{sec:observations}.

\subsection{Prompts with Maximum Rollout Variance are Most Informative}
\label{sec:max_variance_selection}

Here, we show that examples with maximum rollout-reward variance are the most informative for GRPO training. In particular, under the binary-reward setting of GRPO, the magnitude of the gradient signal is directly controlled by the reward variance.


\paragraph{Theorem 1.}
\label{sec:grpo_gradient_bound}
For recent GRPO variants that remove the KL term~\citep{xiao2025dapo}, for a sampled prompt $x$ with a rollout group of size $G$, assume the raw rewards are binary, $r_i \in \{+1, -1\}$, and the gradients of the importance ratios are uniformly bounded as $\|\nabla_\theta \rho_i(\theta)\| \le M$~\citep{razin2025goodteacher}. Then the GRPO objective gradient for this sample satisfies
\begin{equation}
\|\nabla_\theta \mathcal{L}_{\mathrm{GRPO}}\| \le M\sigma.
\end{equation}

\paragraph{Proof sketch.}
The proof proceeds in two steps. We first derive a general upper bound on the GRPO gradient norm by separating the reward-driven term from the KL term and using the uniform bound $\|\nabla_\theta \rho_i(\theta)\| \le M$. 
In the KL-free setting, this reduces the problem to bounding the total absolute advantage signal, $\sum_{i=1}^G |A_i|$. We then analyze the binary-reward case by explicitly writing out the normalized advantages of positive and negative samples and summing their absolute values, which shows that $\sum_{i=1}^G |A_i| = G\sigma$. Substituting this identity into the first bound gives the theorem.  The full proof is in Appendix~\ref{app:proof_theorem1}.
\hfill $\blacksquare$

Theorem 1 shows that the maximum achievable reward-driven gradient signal grows linearly with the within-group reward standard deviation, and is maximized when the binary rewards are perfectly balanced, i.e., $\sigma^2 = 1$.

\subsection{VarIance Guided Online Rollout allocation (VIGOR)}
\label{sec:iterative_rollout_generation}

Motivated by the above observation, we now introduce our method VIGOR. Given a sampled batch $\mathcal{B}$, we initialize the active subset as $\mathcal{A}_1 = \mathcal{B}$. For each prompt $x \in \mathcal{B}$, we initialize an empty rollout set $\mathcal{Y}_x^{(0)} = \emptyset$ and an initial rollout budget $m_x^{(1)} = m_0$. Instead of allocating a large number of rollouts to every prompt upfront, our method performs rollout generation and prompt selection jointly over multiple refinement iterations.

At refinement iteration $t$, let $\mathcal{A}_t$ denote the current active subset of prompts. For each $x \in \mathcal{A}_t$, we first generate $m_x^{(t)}$ rollouts from the current policy $\pi_\theta(\cdot \mid x)$, denote the newly generated rollout set by $\widehat{\mathcal{Y}}_x^{(t)}$, and merge it into the existing rollout set:
\[
\mathcal{Y}_x^{(t)} = \mathcal{Y}_x^{(t-1)} \cup \widehat{\mathcal{Y}}_x^{(t)}.
\]
We then compute the utility score of each prompt in $\mathcal{A}_t$ as the variance of the rewards of its rollouts:
\[
u_x^{(t)} = \mathrm{Var}(\{R(x,y) \mid y \in \mathcal{Y}_x^{(t)}\}),
\qquad x \in \mathcal{A}_t,
\]
and, when $t < T$, construct the next active subset by retaining the top-$\alpha$ fraction of prompts ranked by $u_x^{(t)}$:
\[
\mathcal{A}_{t+1} = \operatorname{Top}_{\alpha}(\mathcal{A}_t; u^{(t)}).
\]
For each retained prompt $x \in \mathcal{A}_{t+1}$, we increase its rollout budget multiplicatively for the next iteration:
\[
m_x^{(t+1)} = \gamma m_x^{(t)},
\]
where $\gamma > 1$ is the expansion ratio. Prompts that are filtered out from $\mathcal{A}_t$ receive no further rollout generation in subsequent iterations of the current training step. By progressively enlarging rollout budgets only for retained prompts, the method concentrates computation on potentially informative examples while avoiding excessive rollout generation on low-value ones. The full procedure is summarized in Algorithm~\ref{alg:iterative_rollout_selection}.
\begin{figure}[t]
\centering
\begin{minipage}{0.85\linewidth}
\begin{algorithm}[H]
\caption{VarIance Guided Online Rollout allocation (VIGOR)}
\label{alg:iterative_rollout_selection}
\KwIn{training dataset $\mathcal{D}$, reward function $R$, total training steps $S$, batch size $B$, initial rollout budget $m_0$, refinement iterations $T$, expansion ratio $\gamma$, selection ratio $\alpha$}
\For{$s = 1$ \KwTo $S$}{
    Sample a batch $\mathcal{B}$ from $\mathcal{D}$ with $|\mathcal{B}| = B$\;
    \ForEach{$x \in \mathcal{B}$}{
        Initialize rollout set $\mathcal{Y}_x \leftarrow \emptyset$ and rollout budget $m_x \leftarrow m_0$\;
    }
    Initialize active subset $\mathcal{A}_1 \leftarrow \mathcal{B}$\;
    \For{$t = 1$ \KwTo $T$}{
        \ForEach{$x \in \mathcal{A}_t$}{
            Generate $m_x$ rollouts from $\pi_\theta(\cdot \mid x)$ and add them to $\mathcal{Y}_x$\;
            Compute utility score $u_x \leftarrow \mathrm{Var}(\{R(x,y) \mid y \in \mathcal{Y}_x\})$\;
        }
        \If{$t < T$}{
            Select a subset $\mathcal{A}_{t+1} \subseteq \mathcal{A}_t$ containing the top-$\alpha$ fraction of prompts ranked by $u_x$\;
            \ForEach{$x \in \mathcal{A}_{t+1}$}{
                Update rollout budget $m_x \leftarrow \gamma m_x$\;
            }
        }
    }
    Use all retained rollouts in $\{\mathcal{Y}_x\}_{x \in \mathcal{B}}$ to perform GRPO updates on $\pi_\theta$\;
}
Return updated policy $\pi_\theta$\;
\end{algorithm}
\end{minipage}
\end{figure}



\subsection{Convergence-Time Analysis}
\label{sec:convergence_time}

\paragraph{Theorem 2.}
Let $u$ denote the prompt-level reward variance, and assume that the upper tail of $u$ follows a Pareto distribution with threshold $u_{\min}>0$ and shape parameter $k>1$, i.e.,
$f(u)=k u_{\min}^k/u^{k+1}$ for $u \ge u_{\min}$. Consider VIGOR with $T$ refinement rounds, selection ratio $\alpha \in (0,1)$, expansion ratio $\gamma>1$, and initial rollout budget $m_0$. In the budget-conserving case $\alpha\gamma=1$, the GRPO-to-VIGOR characteristic timescale ratio is
\begin{equation}
\frac{\tau_{\mathrm{GRPO}}}{\tau_{\mathrm{VIGOR}}}
= \left(
\frac{(\alpha^{-1/k})^T - 1}{T(\alpha^{-1/k} - 1)}
\right)^{1/3}.
\label{eq:timescale_ratio}
\end{equation}
In the compute-skew case $\alpha\gamma>1$, the leading-order scaling becomes
\begin{equation}
\frac{\tau_{\mathrm{GRPO}}}{\tau_{\mathrm{VIGOR}}}
=
\Theta\left((\alpha^{-1/k})^{T/3}\right),
\label{eq:timescale_ratio_compute_skew}
\end{equation}
up to a multiplicative constant depending only on $\alpha$, $\gamma$, and $k$. Thus, under this distributional assumption, VIGOR's timescale advantage grows with the number of refinement rounds and becomes larger for heavier-tailed prompt-variance distributions. The full derivation is given in Appendix~\ref{app:variance_amplification}.

%% file: sec/06_experiments.tex
\section{Experiments}
\label{sec:experiments}

\begin{table*}[t]
\centering
\small
\setlength{\tabcolsep}{5pt}
\caption{Detailed evaluation results on six math reasoning benchmarks. The best and second best performance across all settings are \textbf{bold} and \underline{underscored}. Even with the highly unfair GRESO-g comparison, VIGOR achieves the best average score in nearly all settings.}
\label{tab:main}
\begin{tabular}{lccccccc}
\toprule
\textbf{Method} & \textbf{Math500} & \textbf{AIME24} & \textbf{AMC} & \textbf{Gaokao} & \textbf{Miner.} & \textbf{Olymp.} & \textbf{Avg.} \\
\midrule\midrule
\multicolumn{8}{c}{\textit{Qwen2.5-1.5B}} \\
\midrule
GRPO     & 55.2 & \underline{3.3} & \underline{28.9} & 49.6 & \textbf{20.2} & 19.4 & 29.5 \\
GRESO-r  & 53.0 & \textbf{6.7} & 24.1 & 46.0 & \underline{19.1} & 21.8 & 28.4 \\
GRESO-g  & \textbf{57.8} & \underline{3.3} & \underline{28.9} & \textbf{51.4} & 18.8 & \textbf{24.2} & \textbf{30.7} \\
RL-ZVP   & 55.6 & \underline{3.3} & \textbf{33.7} & 45.7 & 18.8 & 21.5 & 29.8 \\
VIGOR    & \underline{55.8} & \textbf{6.7} & 27.7 & \underline{50.4} & 17.6 & \underline{22.3} & \underline{30.1} \\
\midrule\midrule
\multicolumn{8}{c}{\textit{Qwen2.5-3B}} \\
\midrule
GRPO     & 66.2 & 4.2 & \underline{37.0} & 55.3 & 26.6 & 30.3 & 36.6 \\
GRESO-r  & \textbf{66.6} & 5.0 & 33.7 & \underline{57.1} & \textbf{27.5} & 30.1 & 36.7 \\
GRESO-g  & \underline{66.4} & \underline{8.3} & \underline{37.0} & 56.2 & \underline{27.4} & \textbf{30.7} & \underline{37.7} \\
RL-ZVP   & 65.8 & \textbf{9.2} & 33.7 & \textbf{57.7} & 27.1 & 30.4 & 37.3 \\
VIGOR    & 65.4 & \underline{8.3} & \textbf{39.2} & 56.4 & \textbf{27.5} & \underline{30.6} & \textbf{37.9} \\
\midrule\midrule
\multicolumn{8}{c}{\textit{Qwen2.5-7B}} \\
\midrule
GRPO     & \underline{79.6} & \textbf{23.3} & \underline{55.4} & 64.7 & 33.8 & 41.8 & 49.8 \\
GRESO-r  & 78.2 & 16.7 & 54.2 & 65.7 & \textbf{36.8} & \underline{42.4} & 49.0 \\
GRESO-g  & 79.4 & \underline{20.0} & \textbf{57.8} & \underline{68.1} & 36.0 & 40.7 & \underline{50.3} \\
RL-ZVP   & 77.4 & 16.7 & 54.2 & 66.8 & \underline{36.4} & \textbf{42.6} & 49.0 \\
VIGOR    & \textbf{80.4} & \textbf{23.3} & \underline{55.4} & \textbf{69.1} & 35.3 & 42.3 & \textbf{51.0} \\
\midrule\midrule
\multicolumn{8}{c}{\textit{Phi-4-Mini-Instruct}} \\
\midrule
GRPO     & \textbf{73.6} & \underline{11.7} & 40.1 & \textbf{62.3} & 33.5 & 36.4 & 42.9 \\
GRESO-r  & 72.0 & 9.2 & 40.4 & 59.7 & \underline{33.6} & 35.9 & 41.8 \\
GRESO-g  & \underline{73.4} & \textbf{14.2} & 39.5 & \underline{61.5} & 33.1 & \textbf{37.2} & \underline{43.1} \\
RL-ZVP   & 72.3 & 10.8 & \underline{40.7} & 60.6 & 33.3 & \underline{36.8} & 42.4 \\
VIGOR    & 73.0 & \textbf{14.2} & \textbf{41.6} & \underline{61.5} & \textbf{33.7} & 36.4 & \textbf{43.4} \\
\bottomrule
\end{tabular}
\end{table*}

\subsection{Setup}
\label{sec:exp_setup}
In this section, we present the detailed experimental setup and analyze the experimental results of VIGOR.
\begin{itemize}[leftmargin=15pt]
\item In Section~\ref{sec:exp_main_results}, we show that VIGOR substantially improves training efficiency on both math and coding tasks. On math, VIGOR requires up to 2.3$\times$ fewer rollouts to reach the target accuracy. On coding, VIGOR reaches the same full pass rate with 1.49$\times$ fewer rollouts and improves average test pass rate by 3.4 points. We further show that these efficiency gains come with stronger final performance in both settings.
	\item In Section~\ref{sec:exp_analysis}, we further analyze why VIGOR is effective and validate the effectiveness of variance-based selection through ablation.
\end{itemize}

\paragraph{Models and training configuration.}
We evaluate VIGOR on both mathematical reasoning and coding. For math, we train Qwen2.5-1.5B/3B/7B~\citep{yang2024qwen25} and Phi-4-Mini-Instruct~\citep{phi4mini2025} on MATH~\citep{hendrycks2021math} (mixed with DAPO data~\citep{xiao2025dapo} for the larger models) and report the average accuracy over six reasoning benchmarks. For coding, we follow the rich-feedback LiveCodeBench setting of SDPO~\citep{hubotter2026sdpo} and train Qwen3-8B~\citep{yang2025qwen3} on LiveCodeBench v6 (LCBv6)~\citep{jain2025livecodebench}, reporting full pass rate averaged over three random seeds. For fair comparison, all methods share the same effective batch size, rollout budget per step, and optimizer configuration: VIGOR uses $T=4$, $m_0=2$, $\gamma=2$, and $\alpha=0.5$, while baselines use 8 rollouts per prompt. Full configurations are provided in Appendix~\ref{app:details}.

\paragraph{Baselines.}
We compare VIGOR with the standard GRPO baseline and two recent state-of-the-art methods that represent the two major directions for improving RLVR efficiency: GRESO~\citep{zheng2025greso}, a dynamic-sampling difficulty-aware method~\citep{xiao2025dapo} that uses early-training reward dynamics to sample difficult prompts but requires more rollouts per update, and RL-ZVP~\citep{zvp2026}, which extracts more learning signal from a fixed rollout budget by replacing the zero-advantage signal of zero-variance prompts with entropy-guided token-level rewards. Since GRESO spends extra rollouts, we report two variants in Table~\ref{tab:main}: GRESO-r restricts GRESO to the same total rollout budget as the other methods, while GRESO-g matches their number of gradient-update steps and thus spends about twice as many rollouts. For coding, due to the substantially higher time and compute cost of training on LiveCodeBench v6, we restrict the comparison to GRPO and VIGOR. As a supplementary experiment, we further compare VIGOR with standard GRPO and PODS~\citep{xu2025pods} under a larger rollout budget of $n=32$ (Section~\ref{sec:exp_analysis}).

\subsection{Main Results}
\label{sec:exp_main_results}

\paragraph{Main math results.}
We first evaluate VIGOR on mathematical reasoning. The right panel of Figure~\ref{fig:score_vs_rollouts} shows that VIGOR consistently reduces the rollout budget needed to reach the target accuracy. The largest improvement appears on Qwen2.5-3B, where VIGOR reaches the target with 2.3$\times$ fewer rollouts than GRPO, GRESO, and RL-ZVP. We defer the per-model rollout-efficiency curves for the remaining scales to Figure~\ref{fig:score_vs_rollouts_additional_math} in Appendix~\ref{app:detailed_benchmark}, which show the same advantage on Qwen2.5-1.5B, Qwen2.5-7B, and Phi-4-Mini-Instruct.

Table~\ref{tab:main} reports the detailed benchmark scores after training, using two epochs for the Qwen models and one epoch for Phi-4-Mini-Instruct, and selecting each method's best average-score checkpoint. Under the fair rollout-matched comparison against GRPO, GRESO-r, and RL-ZVP, VIGOR achieves the best average score on all four model scales. It also improves the strongest competing method on key benchmarks, with gains of up to 2.5 points on the best benchmark. Even compared with GRESO-g, which uses about twice the rollout computation, VIGOR still achieves a higher average score on all but the 1.5B model. These results show that VIGOR improves rollout efficiency while preserving, and often improving, final model quality.

\paragraph{Coding results on LiveCodeBench v6.}
The left panel of Figure~\ref{fig:score_vs_rollouts} further shows that VIGOR improves RLVR training on LCBv6 with Qwen3-8B. In this figure, full pass rate counts a problem as correct only if all of its validation tests pass. VIGOR reaches the full-pass-rate level achieved by GRPO at the end of training using 32.9\% fewer rollouts, corresponding to a 1.49$\times$ rollout-efficiency speedup. At the end of training, VIGOR also improves the three-seed average full pass rate from 44.0\% to 45.7\%, a gain of 1.7 percentage points over GRPO. In Appendix~\ref{app:multiseed_robustness}, we further report the finer-grained average test pass rate, which averages the pass rate over individual test cases rather than requiring a full solve: VIGOR improves this metric from 63.4\% to 66.8\% at the end of training, a gain of 3.4 percentage points. These coding results show that VIGOR remains effective beyond math reasoning and continues to improve rollout efficiency and final performance on more complex program-synthesis tasks.


\begin{figure}[h]
	\centering
	\begin{minipage}[t]{0.4\textwidth}
	\includegraphics[width=\linewidth]{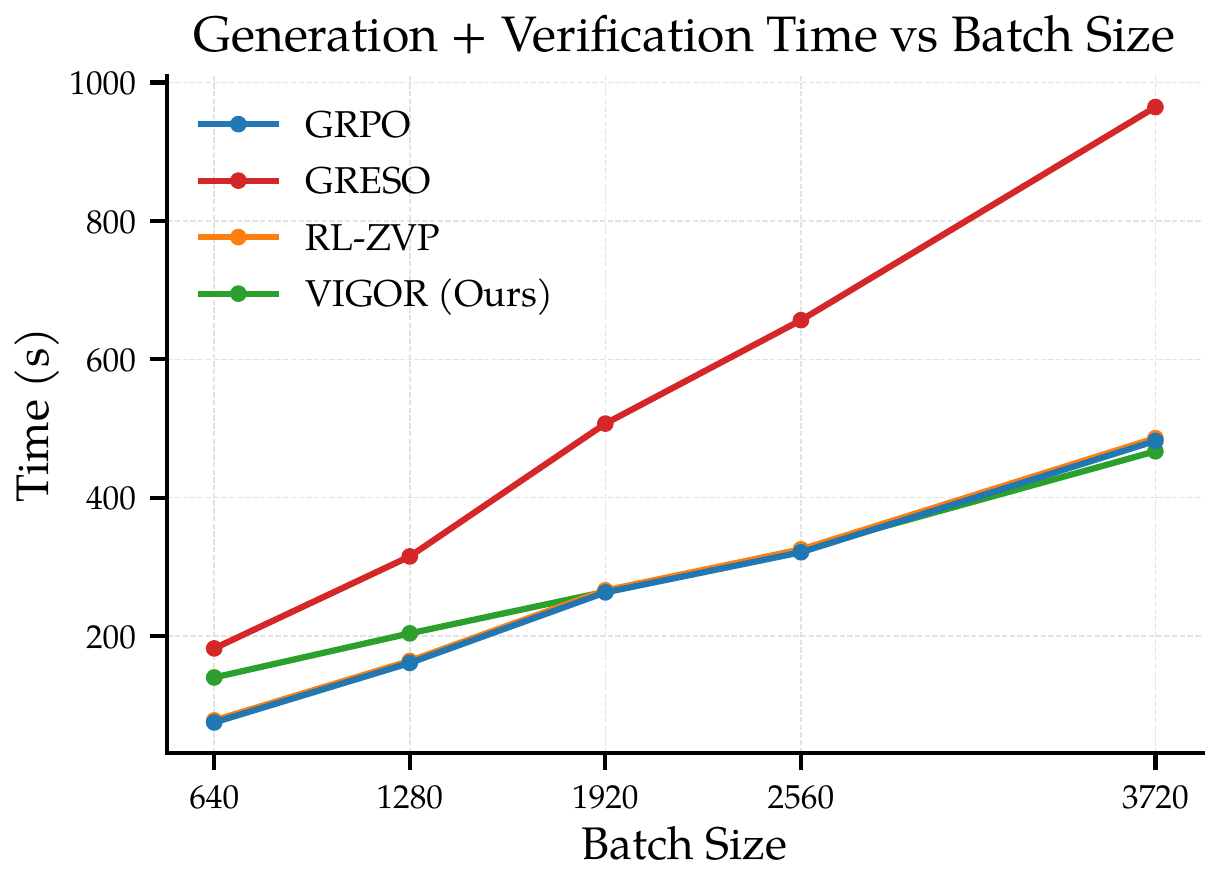}
	\end{minipage}
	\hspace{0.03\textwidth}
	\begin{minipage}[t]{0.4\textwidth}
	\includegraphics[width=\linewidth]{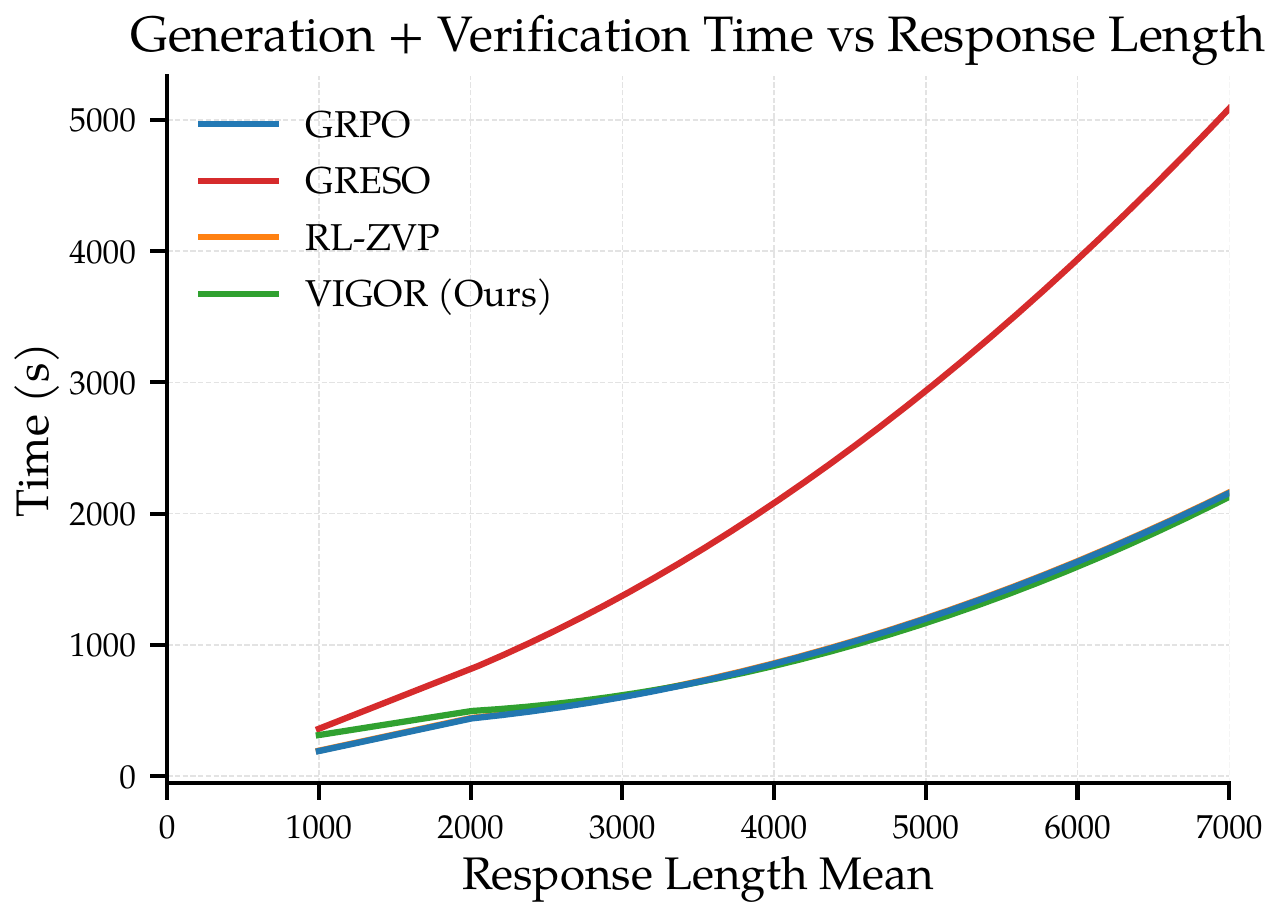}
	\end{minipage}
	\vspace{-2mm}
	\caption{Generation and verification time under different scaling factors, measured on 8$\times$GH200 GPUs. VIGOR achieves substantial training speedups over GRPO without incurring additional time, and requires far less time than GRESO. \textbf{Left}: Time versus batch size on Qwen2.5-7B. \textbf{Right}: Time versus average response length at fixed batch size 640 on Phi-4-mini-instruct and Phi-4-mini-reasoning.\looseness=-1}
	\label{fig:timing_analysis}
	\end{figure}
\paragraph{No extra time overhead at the same rollout budget.}
All methods use the same number of prompts, gradient updates, and rollout budget per step, except GRESO, which requires more rollouts. We measure wall-clock generation-and-verification time under varying batch sizes and response lengths, as shown in Figure~\ref{fig:timing_analysis}. VIGOR stays close to the baseline and RL-ZVP, and becomes substantially faster than GRESO once the batch size is sufficiently large. Since this time grows superlinearly with response length, GRESO's overhead is especially pronounced on today's long-chain-of-thought workloads, whereas VIGOR preserves near-baseline efficiency despite its iterative rollout process.

This also justifies using rollout count as a direct efficiency proxy in the main curves of Figure~\ref{fig:score_vs_rollouts}: the gradient-update phase is nearly identical across methods, while generation time scales almost linearly with rollout count once the batch is large enough to saturate the rollout engine. Combined with convergence speed, this yields an end-to-end wall-clock advantage: on Qwen2.5-3B, VIGOR reaches the 37.8\% target accuracy in 50 steps versus GRESO's 105, a $2.65\times$ wall-clock speedup.

\subsection{Analysis and Ablation Study}
\label{sec:exp_analysis}

\begin{figure}[h]
\centering
\begin{minipage}[t]{0.4\textwidth}
\includegraphics[width=\linewidth]{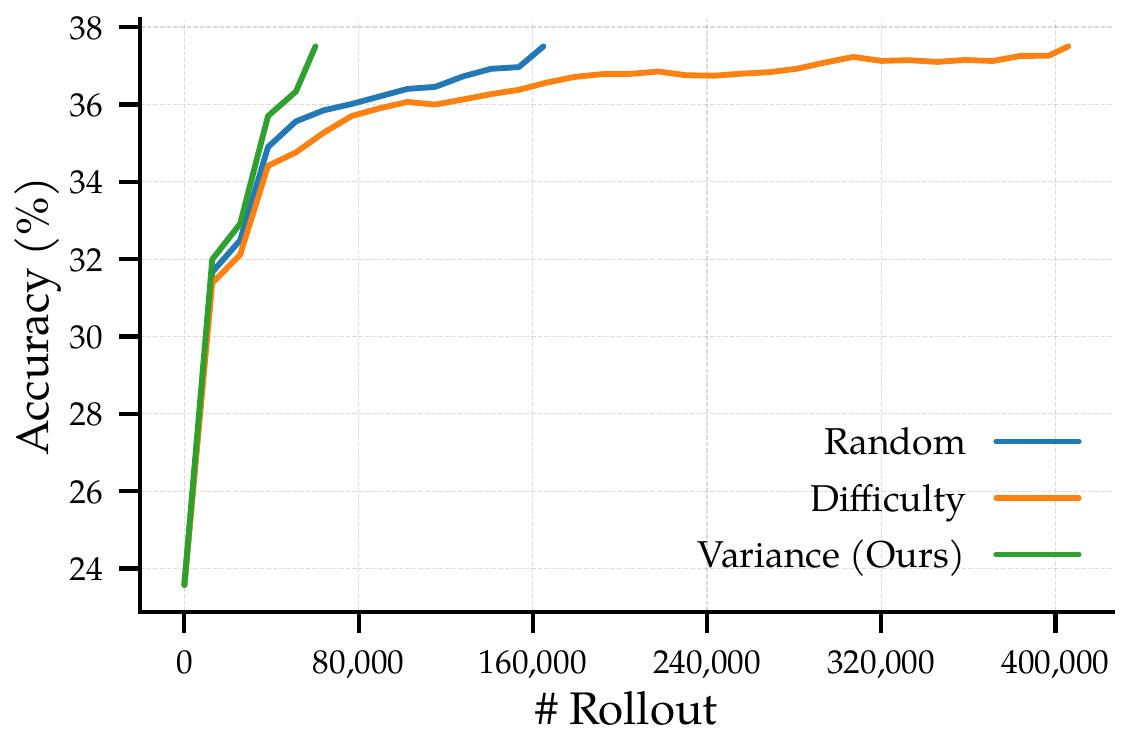}
\end{minipage}
\hspace{0.02\textwidth}
\begin{minipage}[t]{0.4\textwidth}
\centering
\includegraphics[width=\linewidth]{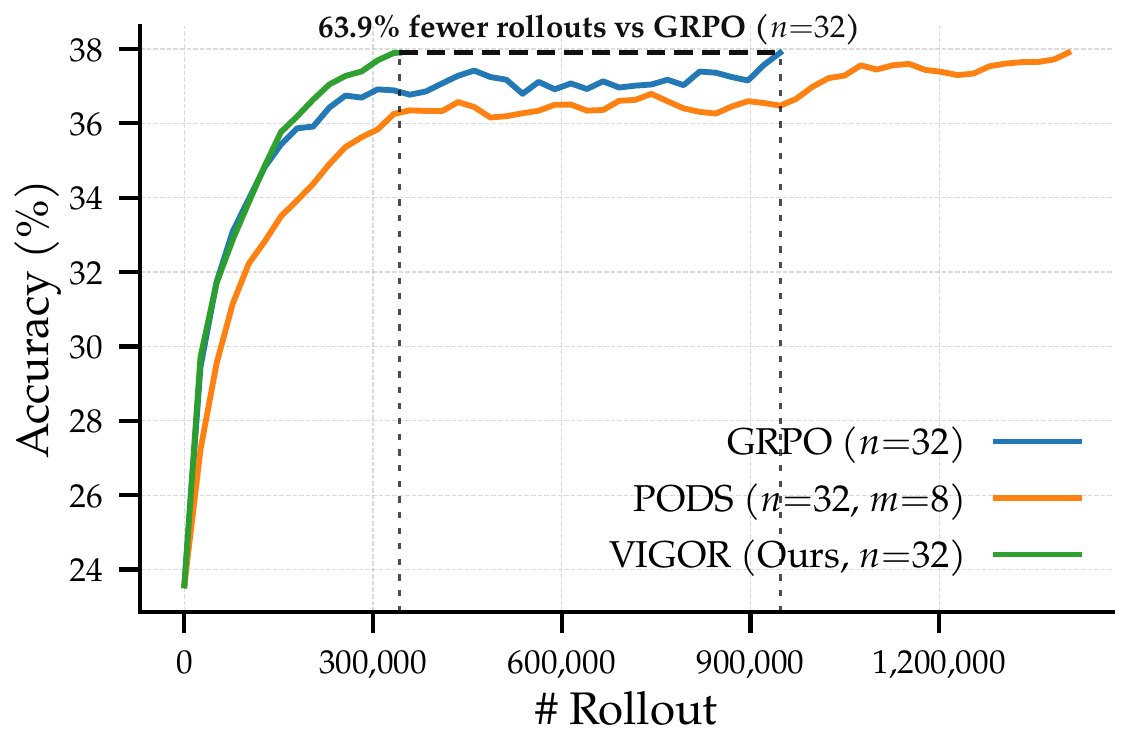}
\end{minipage}
\vspace{-2mm}
\caption{Ablation and large-group analysis. \textbf{Left}: Ablation on prompt selection strategies. We compare our variance-based selection (VIGOR) with difficulty-based and random selection on Qwen2.5-3B under the same training setting. VIGOR consistently achieves the best performance, supporting our theoretical motivation. \textbf{Right}: Large-group comparison under $n=32$, where VIGOR remains more rollout-efficient than GRPO and PODS.\looseness=-1}
\label{fig:ablation_large_group}
\end{figure}

\paragraph{Variance-based selection achieves faster convergence with fewer rollouts.}
The core design choice of VIGOR is to allocate additional rollout budget according to within-group reward variance. To validate this choice, we compare our variance-based selection (VIGOR) against difficulty-based and random selection in the ablation study. As shown in the left panel of Figure~\ref{fig:ablation_large_group}, VIGOR consistently achieves the best performance, confirming that its gains come from prioritizing more informative prompts rather than from generic computation reallocation alone. This suggests that selecting samples with larger gradient magnitude matters more than merely reducing ineffective samples with zero reward variance.

\paragraph{VIGOR remains effective at large rollout budgets.}
We further compare VIGOR with standard GRPO and PODS~\citep{xu2025pods} under a larger candidate group size of $n=32$. As shown in the right panel of Figure~\ref{fig:ablation_large_group}, VIGOR still achieves a $2.77\times$ rollout-efficiency speedup over GRPO, showing that the benefit is not limited to small group sizes with noisy variance estimates. Compared with PODS, which generates a large candidate pool and then down-samples rollouts before the policy update, VIGOR performs allocation during generation and avoids spending rollout budget on candidates that will later be discarded. Under the matched-budget setting, PODS requires $5.34\times$ more rollouts and $3.97\times$ more wall-clock time than VIGOR to reach the same target performance.

In Appendix~\ref{app:more_ablations} and Appendix~\ref{app:sample_ratio_curriculum}, we further provide a hyperparameter ablation study and additional analyses on VIGOR's effective sample ratio and emergent training curriculum.


%% file: sec/07_conclusion.tex
\section{Conclusion}
\label{sec:conclusion}
\vspace{-2mm}
In this paper, we proposed VIGOR, an iterative data selection framework for RLVR, in particular GRPO. 
By progressively allocating more rollout budget only to examples with larger within-group reward variance, VIGOR improves computational efficiency while preserving the original GRPO training paradigm. Experiments on mathematical reasoning and coding tasks show that VIGOR substantially reduces the rollout budget needed to reach the same performance, achieving up to 2.3$\times$ fewer rollouts on math and 1.49$\times$ fewer rollouts to reach GRPO's final full pass rate on coding, while also improving the average coding test pass rate by 3.4 points. These results suggest that variance-guided iterative rollout generation is a promising direction for more compute-efficient RLVR.

%% file: sec/appendix.tex
\section{Detailed Benchmark Results}
\label{app:detailed_benchmark}

\begin{figure}[h]
\centering
\includegraphics[width=\linewidth]{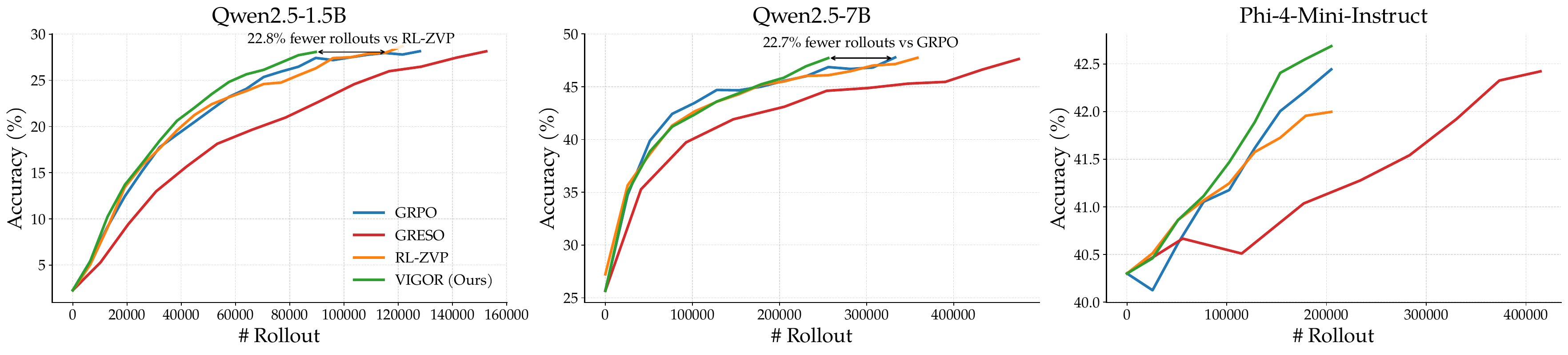}
\caption{Rollout efficiency on the remaining math model settings. VIGOR is compared with GRPO, GRESO, and RL-ZVP on Qwen2.5-1.5B, Qwen2.5-7B, and Phi-4-Mini-Instruct.}
\label{fig:score_vs_rollouts_additional_math}
\end{figure}

We provide the validation-accuracy curves against rollout compute over the course of training for all the remaining model settings in Figure~\ref{fig:score_vs_rollouts_additional_math}. Unlike the main-paper summary views, these curves expose the entire training process and make it possible to compare both the learning speed and the final performance of different data selection strategies throughout optimization.

Across models, our method achieves more efficient data selection over the course of training, which translates into faster accuracy improvements under the same rollout budget. In particular, the gains are already visible in the early and middle stages of training, and the advantage is maintained as training proceeds, indicating that the selected data remains informative beyond only the initial warm-up phase.

\section{More Ablations}
\label{app:more_ablations}


As shown in the left panel of Figure~\ref{fig:sample_ratio_curriculum}, even with VIGOR, roughly $20\%$ of the samples at each gradient update step are still invalid rollouts. This observation motivates a natural follow-up question: should we further rebalance the rollout budget by using more rollouts per iteration but fewer iterative refinement rounds, or by relaxing the selection ratio after the first generation stage so that more prompts remain active for later rollout allocation?

\begin{wrapfigure}{r}{0.5\textwidth}
\centering
\vspace{-0.8em}
\includegraphics[width=\linewidth]{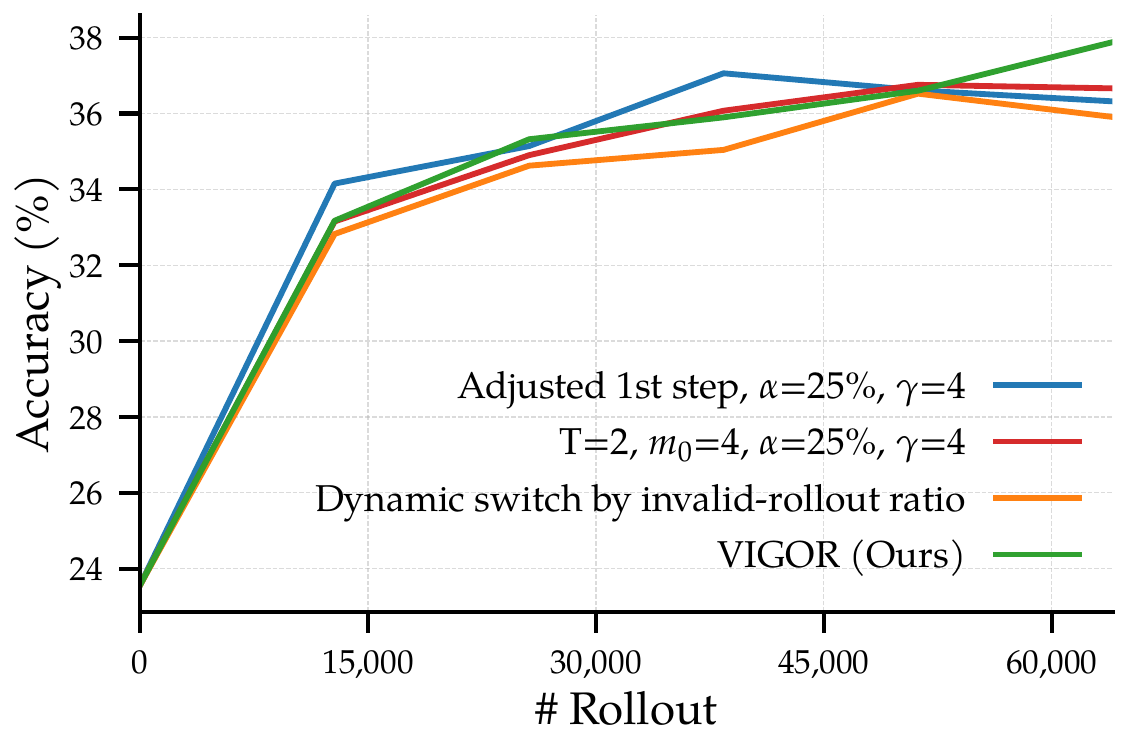}
\caption{Ablation on alternative rollout-allocation schedules measured by validation accuracy against the number of rollouts consumed. Every method uses the same total number of rollouts per gradient update step for a fair comparison.}
\label{fig:appendix-valsel-ablation}
\vspace{-0.8em}
\end{wrapfigure}

To study this trade-off, we compare VIGOR with three alternative designs. The first variant, \emph{T=2, $m_0$=4, $\alpha$=25\%, $\gamma$=4}, increases the number of rollouts generated in each iteration while reducing the number of iterative refinement rounds. The second variant, \emph{Adjusted 1st step, $\alpha$=25\%, $\gamma$=4}, keeps more prompts after the first generation stage by lowering the pruning ratio, and then compensates with more rollout allocation in subsequent refinement. The third variant, \emph{Dynamic switch by invalid-rollout ratio}, adaptively changes the rollout schedule according to the observed invalid-rollout ratio during training. In all cases, we keep the total rollout budget per gradient update step identical, so the comparison isolates the effect of allocation strategy rather than total computation.

Figure~\ref{fig:appendix-valsel-ablation} shows that, overall, the default VIGOR configuration achieves the best performance. The \emph{Adjusted 1st step, $\alpha$=25\%, $\gamma$=4} variant shows a weak advantage during the first refinement iteration, but this early gain is overtaken as training progresses, and VIGOR ultimately reaches the highest final accuracy; we further find that this early-stage improvement is closely tied to the specific dataset and does not generalize across settings. This suggests that simply reducing the invalid-rollout ratio is not sufficient: the key is to preserve iterative selection pressure on high-variance prompts, which allows VIGOR to allocate computation to samples with stronger learning value throughout training.

All schedules in this ablation are constrained to the same total rollout budget per gradient update step. The default configuration, $T=4$, $m_0=2$, $\gamma=2$, and $\alpha=0.5$, therefore reflects an allocation trade-off rather than a larger compute budget: it retains enough rounds to repeatedly focus on high-variance prompts while keeping the first-stage estimate inexpensive.

\section{Detailed Experimental Setup}
\label{app:details}

We present the full experimental setup details here. VIGOR and all baselines, including GRPO, GRESO, and RL-ZVP, are implemented within the same verl-based training pipeline~\citep{sheng2024hybridflow} with vLLM~\citep{kwon2023vllm} for rollout generation, ensuring that every comparison is conducted under a unified infrastructure. For fair comparison, all methods use the same effective training batch size and optimizer configuration. VIGOR uses $T=4$, $m_0=2$, $\gamma=2$, and $\alpha=0.5$, while all baselines use 8 rollouts per prompt so that the total rollout budget per step is matched under the same batch size. We conduct math experiments on Qwen2.5-1.5B, Qwen2.5-3B, Qwen2.5-7B~\citep{yang2024qwen25}, and Phi-4-Mini-Instruct~\citep{phi4mini2025}. For the 1.5B and 3B models, training uses only the MATH dataset, while for the 7B and Phi models we follow prior work~\citep{zheng2025greso} and train on a mixture of MATH and DAPO data to increase data diversity at the larger scale. For Qwen2.5-1.5B/3B/7B, we use a maximum context length of 4096 tokens, split into a maximum prompt length of 1536 and a maximum response length of 2560. For Phi-4-Mini-Instruct, we instead use a maximum prompt length of 2048 and a maximum response length of 6144. For reward computation, we use Math-Verify~\citep{kydlicek2024mathverify}. The main training runs are conducted on 8$\times$GH200 GPUs, while several statistics-related experiments are conducted on 4$\times$A40 GPUs. The training batch size is 160 for the 1.5B and 3B models and 640 for the 7B and Phi-4-Mini-Instruct models, with gradient-update mini-batch sizes of 32 and 64, respectively; each training step contains 20 gradient-update steps. We optimize the actor model with AdamW~\citep{loshchilov2019adamw} using a constant learning rate of $1\times10^{-6}$. Unless otherwise stated, the rollout engine memory utilization is capped at 0.5, the reference model log-probability micro-batch size is 32 per GPU with parameter offloading enabled, and critic warmup is set to 0. For the baseline methods, we follow the open-source GRESO implementation exactly, while for RL-ZVP we follow the paper setting and use a scaling factor of $\alpha=0.1$.

For evaluation, we run validation every 5 training steps, corresponding to every 100 gradient-update steps, to measure training efficiency throughout optimization. During validation, we generate one sampled response per prompt with top-$p=0.7$ and temperature $1.0$. We report results on six mathematical reasoning benchmarks: Math500~\citep{lightman2023lets}, AIME24~\citep{aops2024aime24}, AMC~\citep{aops2024amc}, Minerva Math~\citep{lewkowycz2022minerva}, Gaokao~\citep{zhang2023gaokao}, and Olympiad Bench~\citep{he2024olympiadbench}. Following the main experimental protocol, the results in Table~\ref{tab:main} are taken from the best checkpoint of each method after training for three epochs on MATH for the 1.5B and 3B models, and for two epochs on the MATH+DAPO mixture for the 7B model.

\paragraph{Coding setup.}
For coding, we follow the rich-feedback LiveCodeBench setting of SDPO~\citep{hubotter2026sdpo} and train Qwen3-8B~\citep{yang2025qwen3} on LiveCodeBench v6 (LCBv6)~\citep{jain2025livecodebench}. LCBv6 contains recent contest-style programming problems and naturally exposes informative environment feedback, such as public unit-test outcomes, failed test cases, and runtime errors. We evaluate under a public/private unit-test protocol: public tests provide online training feedback, while hidden/private tests are reserved for validation. During training, each generated solution is executed against the public tests, yielding a verifiable pass/fail reward together with rich execution feedback. VIGOR keeps the GRPO optimization objective unchanged and uses the emerging group reward variance to decide which coding prompts receive additional rollouts. We use a maximum prompt length of 2048 and a maximum response length of 8192. The training batch size is 64, with a PPO mini-batch size of 8 and a micro-batch size of 1 per GPU; we disable the KL loss, following the KL-free GRPO setup used throughout this work, and use an asymmetric clipping ratio with an upper bound of 0.28. Training is conducted on 16 GPUs, with the rollout engine memory utilization capped at 0.55. As in the math setting, VIGOR starts from an initial rollout budget of $m_0=2$ per prompt. We validate every 5 training steps, sampling 4 responses per prompt with top-$p=0.95$ and temperature $0.6$, and train for 31 epochs in total. Due to the substantially longer training time and stochasticity of coding RL, we compare VIGOR with the standard GRPO baseline under the same Qwen3-8B and LCBv6 setup, and run each method with three random seeds. We report full pass rate on the validation tests, averaged across the three seeds.

\section{Proofs}
\label{app:proofs}

In this section, we provide the deferred proofs for the theoretical results in Section~\ref{sec:experiments}.

\subsection{Proof of Theorem 1}
\label{app:proof_theorem1}

\paragraph{Lemma 1.}
For a sampled prompt $x$ with a rollout group of size $G$, assume the importance sampling gradients are bounded, $\|\nabla_\theta \rho_i(\theta)\| \le M$~\citep{razin2025goodteacher}. Define the total advantage signal as $S = \sum_{i=1}^G |A_i|$. Then the gradient of the GRPO objective for this sample is bounded by
\begin{equation}
\|\nabla_\theta \mathcal{L}_{\mathrm{GRPO}}\|
\le
\frac{M}{G} S + \beta \, \|\nabla_\theta D_{\mathrm{KL}}(\pi_\theta \,\|\, \pi_{\mathrm{ref}})\|.
\end{equation}

\paragraph{Proof.}
Let $I_i \in \{0, 1\}$ be the indicator for the unclipped surrogate objective. The objective gradient is
\begin{equation}
\nabla_\theta \mathcal{L}_{\mathrm{GRPO}}
= \frac{1}{G} \sum_{i=1}^G I_i A_i \, \nabla_\theta \rho_i(\theta) - \beta \, \nabla_\theta D_{\mathrm{KL}}(\pi_\theta \,\|\, \pi_{\mathrm{ref}}).
\end{equation}
Applying the triangle inequality and $|I_i| \le 1$ to the reward-driven term yields
\begin{equation}
\begin{aligned}
\left\| \frac{1}{G} \sum_{i=1}^G I_i A_i \, \nabla_\theta \rho_i(\theta) \right\|
&\le
\frac{1}{G} \sum_{i=1}^G |I_i| \cdot |A_i| \cdot \|\nabla_\theta \rho_i(\theta)\| \\
&\le \frac{M}{G} S.
\end{aligned}
\end{equation}
Adding the KL penalty term gives
\begin{equation}
\|\nabla_\theta \mathcal{L}_{\mathrm{GRPO}}\|
\le
\frac{M}{G} S + \beta \, \|\nabla_\theta D_{\mathrm{KL}}(\pi_\theta \,\|\, \pi_{\mathrm{ref}})\|.
\end{equation}
Recent RLVR systems often remove the KL term to improve training efficiency~\citep{xiao2025dapo,zheng2025greso}. In this setting, the bound reduces to
\begin{equation}
\|\nabla_\theta \mathcal{L}_{\mathrm{GRPO}}\| \le \frac{M}{G} S.
\end{equation}
\hfill $\blacksquare$

\vspace{1em}

\paragraph{Lemma 2.}
For a group of $G$ generated samples with binary raw rewards $r_i \in \{+1, -1\}$, define the total gradient signal in GRPO as $S = \sum_{i=1}^G |A_i|$. Then $S$ is proportional to the within-group reward standard deviation with proportionality constant $G$, i.e., $S = G\sigma$.

\paragraph{Proof.}
Let $p$ be the proportion of samples receiving a positive reward $+1$. The mean and variance of the binary rewards are $\mu = 2p - 1$ and $\sigma^2 = 4p(1-p)$, respectively. The z-score normalized advantages, defined as $A_i = \frac{r_i - \mu}{\sigma}$, for the positive and negative samples are
\begin{equation}
 A_+ = \frac{2(1-p)}{\sigma}, \qquad A_- = \frac{-2p}{\sigma}.
\end{equation}
The total gradient signal $S = \sum_{i=1}^G |A_i|$ is the sum of absolute advantages over the $Gp$ positive and $G(1-p)$ negative samples:
\begin{equation}
S = G \left( p |A_+| + (1-p) |A_-| \right) = G \left( \frac{2p(1-p)}{\sigma} + \frac{2p(1-p)}{\sigma} \right) = G \frac{4p(1-p)}{\sigma}.
\end{equation}
Since $\sigma^2 = 4p(1-p)$, this simplifies directly to $S = G \frac{\sigma^2}{\sigma} = G\sigma$. \hfill $\blacksquare$

\paragraph{Proof of Theorem 1.}
Following the KL-free training setup considered in Theorem 1~\citep{xiao2025dapo}, Lemma 1 gives
\begin{equation}
\|\nabla_\theta \mathcal{L}_{\mathrm{GRPO}}\| \le \frac{M}{G} S.
\end{equation}
By Lemma 2, in the binary-reward setting we have $S = G\sigma$. Substituting this identity into the bound above yields
\begin{equation}
\|\nabla_\theta \mathcal{L}_{\mathrm{GRPO}}\| \le M\sigma.
\end{equation}
Therefore, in the binary-reward setting of GRPO, the gradient signal magnitude is directly controlled by the within-group reward dispersion: larger sample variance yields larger $\sigma$, and thus stronger update signal. In particular, the maximum $S = G$ is attained when the rewards are perfectly balanced, i.e., $\sigma^2 = 1$. This proves the theorem. \hfill $\blacksquare$

\subsection{Variance Amplification and Training Timescale}
\label{app:variance_amplification}

We provide the distributional calculation behind Equation~\ref{eq:timescale_ratio}. We first recall the optimization-timescale result of \citet{razin2025goodteacher}. For a desired reward improvement $\gamma>0$, they define the hitting time
\begin{equation}
t_\gamma =
\inf \left\{
t :
\mathbb{E}_{y \sim \pi_{\theta(t)}(\cdot \mid x)}[r(x,y)]
\ge
\mathbb{E}_{y \sim \pi_{\theta(0)}(\cdot \mid x)}[r(x,y)] + \gamma
\right\}.
\end{equation}
They show that this time is lower bounded by
\begin{equation}
\Omega \left(
\left(
\mathbb{E}_{x' \sim \mathcal{S}}
\left[
\operatorname{Var}_{y \sim \pi_{\theta(0)}(\cdot \mid x')}
[r_{RM}(x',y)]
\right]
\right)^{-1/3}
\right),
\end{equation}
where the hidden constants depend on optimization-specific factors such as the target improvement, KL coefficient, and sequence length. Thus, optimization time scales with the inverse cubic root of expected reward variance, and increasing the effective reward variance reduces the characteristic optimization timescale.

We now analyze how VIGOR increases the effective variance. This analysis is intended to characterize the regime where prompt informativeness is strongly heterogeneous, which is common in RLVR: most prompts are already easy or still too hard and therefore have near-zero reward variance, while a small frontier subset has high variance and contributes stronger policy-gradient signal.

Let $u$ denote the prompt-level reward variance. We model the upper tail of $u$ with a Pareto distribution with threshold $u_{\min}>0$ and shape parameter $k>1$:
\begin{equation}
f(u)=\frac{k u_{\min}^k}{u^{k+1}}, \qquad u \ge u_{\min}.
\end{equation}
Smaller $k$ corresponds to a heavier tail, where informative prompts are rarer but have much larger variance.

\paragraph{Uniform GRPO allocation.}
Under uniform prompt sampling, the effective prompt variance is
\begin{equation}
\mathrm{Var}_{\mathrm{GRPO}}
= \mathbb{E}[u]
= \int_{u_{\min}}^\infty u f(u) \,du
= \frac{k u_{\min}}{k-1}.
\end{equation}

\paragraph{VIGOR allocation.}
At refinement round $i$, VIGOR retains the top $\alpha^i$ fraction of prompts and allocates $\gamma^i m_0$ rollouts to each retained prompt. The survival threshold $u_i^\star$ satisfies
\begin{equation}
\mathbb{P}(u>u_i^\star)=\alpha^i,
\qquad
u_i^\star = u_{\min}\alpha^{-i/k}.
\end{equation}
The variance contribution of round $i$ is therefore
\begin{equation}
\begin{aligned}
E_{\mathrm{var}}^{(i)}
&= \gamma^i m_0 \int_{u_i^\star}^{\infty} u f(u)\,du \\
&= m_0 \gamma^i \mathrm{Var}_{\mathrm{GRPO}}
\alpha^{i(1-1/k)}.
\end{aligned}
\end{equation}
The expected rollout count in the same round is
\begin{equation}
E_{\mathrm{count}}^{(i)} = m_0 \gamma^i \alpha^i.
\end{equation}
Thus the exact effective-variance amplification over $T$ rounds is
\begin{equation}
\eta_{\mathrm{exact}}
=
\frac{\mathrm{Var}_{\mathrm{VIGOR}}}{\mathrm{Var}_{\mathrm{GRPO}}}
=
\frac{\sum_{i=0}^{T-1}(\gamma\alpha^{1-1/k})^i}
{\sum_{i=0}^{T-1}(\alpha\gamma)^i}.
\label{eq:eta_exact}
\end{equation}
Since $\gamma\alpha^{1-1/k}=(\alpha\gamma)\alpha^{-1/k}$ and $\alpha^{-1/k}>1$, the numerator grows faster than the denominator. This gap becomes larger as the variance distribution becomes more heavy-tailed.

\paragraph{Timescale reduction.}
Following the result above, we write the characteristic training time as
\begin{equation}
\tau = C \cdot \mathrm{Var}^{-1/3},
\end{equation}
where $C>0$ absorbs optimization-dependent constants. Therefore
\begin{equation}
\frac{\tau_{\mathrm{GRPO}}}{\tau_{\mathrm{VIGOR}}}
= \eta_{\mathrm{exact}}^{1/3}.
\end{equation}
For the compute-skew case $\alpha\gamma>1$, the leading-order scaling is
\begin{equation}
\frac{\tau_{\mathrm{GRPO}}}{\tau_{\mathrm{VIGOR}}}
\simeq
(\alpha^{-1/k})^{T/3} C(\alpha,\gamma,k)^{1/3},
\end{equation}
where $C(\alpha,\gamma,k)$ is independent of $T$. For the budget-conserving case $\alpha\gamma=1$, Equation~\ref{eq:eta_exact} simplifies to
\begin{equation}
\eta_{\alpha\gamma=1}
=
\frac{\sum_{i=0}^{T-1}(\alpha^{-1/k})^i}{T}
=
\frac{(\alpha^{-1/k})^T - 1}{T(\alpha^{-1/k}-1)},
\end{equation}
which gives
\begin{equation}
\frac{\tau_{\mathrm{GRPO}}}{\tau_{\mathrm{VIGOR}}}
=
\left(
\frac{(\alpha^{-1/k})^T - 1}{T(\alpha^{-1/k}-1)}
\right)^{1/3}.
\end{equation}
This expression shows that VIGOR's relative advantage increases as $k \to 1^+$, i.e., as the prompt-variance distribution becomes more heavy-tailed.

\section{Additional Robustness Analyses}
\label{app:additional_robustness}

\subsection{Multi-Seed Robustness}
\label{app:multiseed_robustness}

To assess stability across random seeds, we run multi-seed evaluations in both domains, training each method across three distinct seeds on MATH with Qwen2.5-1.5B and on LCBv6 with Qwen3-8B. As shown in the left and middle panels of Figure~\ref{fig:lcb_multiseed}, VIGOR remains stable across seeds in both settings, achieving an average training-efficiency speedup of 42.1\% over GRPO on math while preserving its rollout-efficiency and final-accuracy advantage on coding. These consistent gains across domains and seeds demonstrate the strong robustness of our method.

\subsection{Initial Variance Estimates Have a Low False-Negative Rate}
\label{app:false_negative_rate}

\begin{figure}[h]
\centering
\begin{minipage}[t]{0.32\textwidth}
\centering
\includegraphics[width=\linewidth]{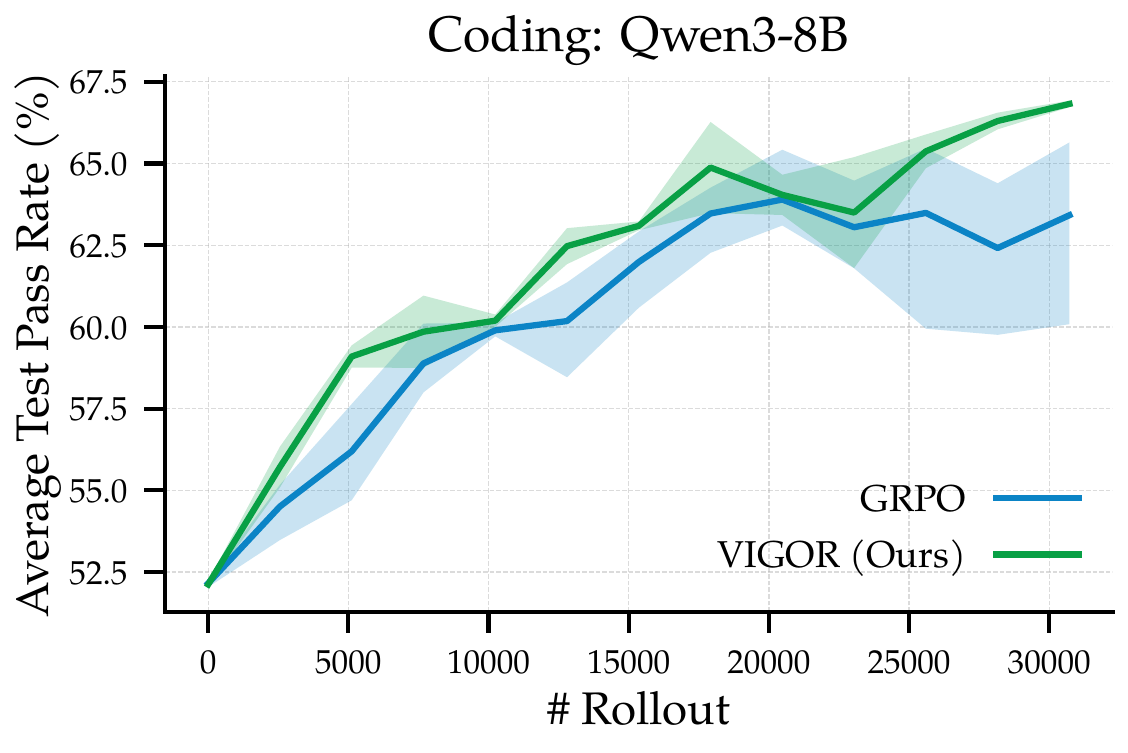}
\end{minipage}
\hfill
\begin{minipage}[t]{0.32\textwidth}
\centering
\includegraphics[width=\linewidth]{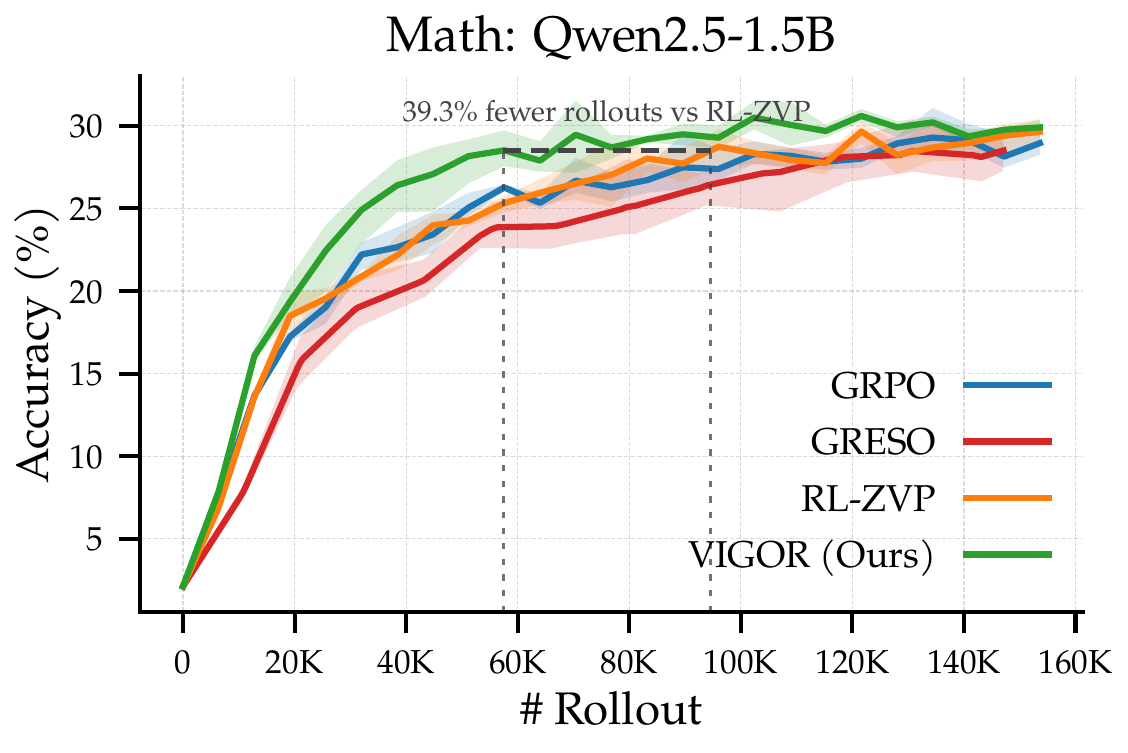}
\end{minipage}
\hfill
\begin{minipage}[t]{0.32\textwidth}
\centering
\includegraphics[width=\linewidth]{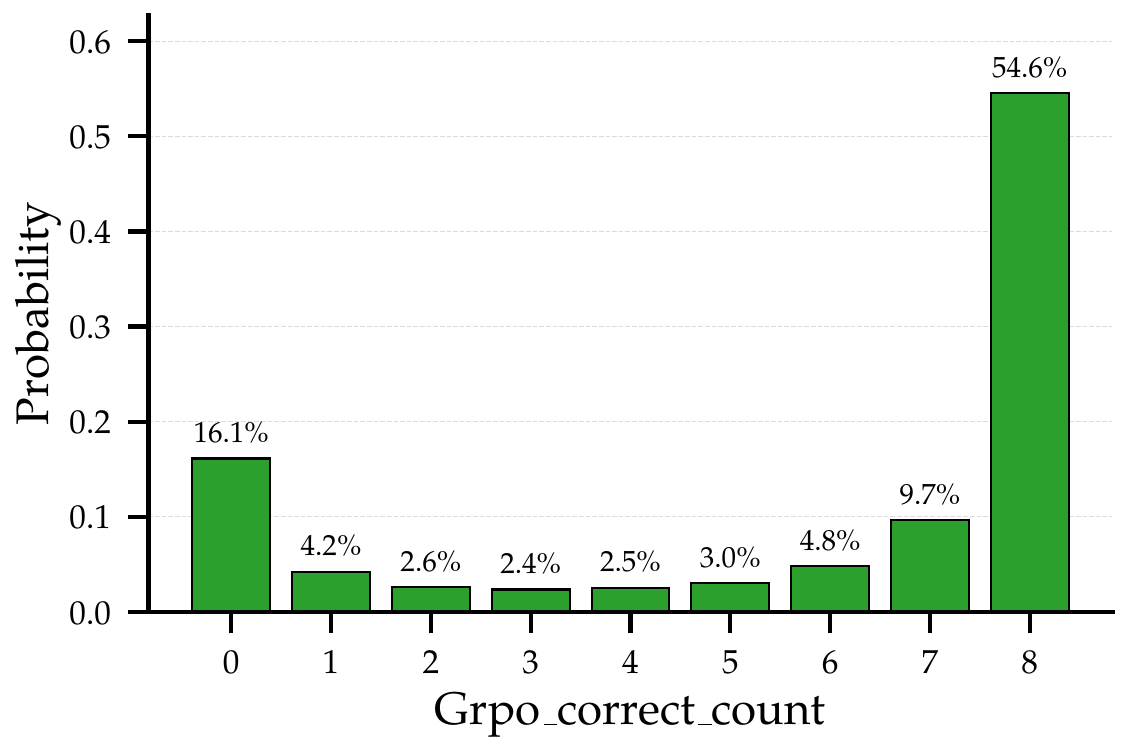}
\end{minipage}
\vspace{-2mm}
\caption{Additional robustness results. \textbf{Left}: average test pass rate on LCBv6 with Qwen3-8B. \textbf{Middle}: multi-seed training curves on MATH with Qwen2.5-1.5B, where each method is run across three random seeds (shaded regions denote variation across seeds) and VIGOR achieves an average training-efficiency speedup of 42.1\% over GRPO. \textbf{Right}: distribution of the number of correct rollouts (out of 8 measurement rollouts) for prompts that VIGOR assigned to the low-budget track after the initial $m_0=2$ rollouts; only $7.9\%$ of these prompts fall in the high-variance frontier region, indicating a low false-negative rate for the initial variance estimate.}
\label{fig:lcb_multiseed}
\end{figure}

Since VIGOR starts with only $m_0=2$ rollouts per prompt, we measure whether informative prompts are often missed by the initial variance estimate. In a Qwen2.5-3B training run on MATH, we additionally generated 8 rollouts per prompt for measurement only, while keeping VIGOR's allocation unchanged. Among prompts assigned to the low-budget track after the initial two rollouts, prompts that would have fallen in the high-variance frontier region under all 8 measurement rollouts account for only 7.9\%. As shown in the right panel of Figure~\ref{fig:lcb_multiseed}, this indicates that the initial estimate filters out mostly low-information prompts, and any missed frontier prompts can still reappear in later epochs as the policy changes.

\section{Effective Sample Ratio and Emergent Curriculum}
\label{app:sample_ratio_curriculum}

\begin{figure}[h]
\centering
\begin{minipage}[t]{0.4\textwidth}
\includegraphics[width=\linewidth]{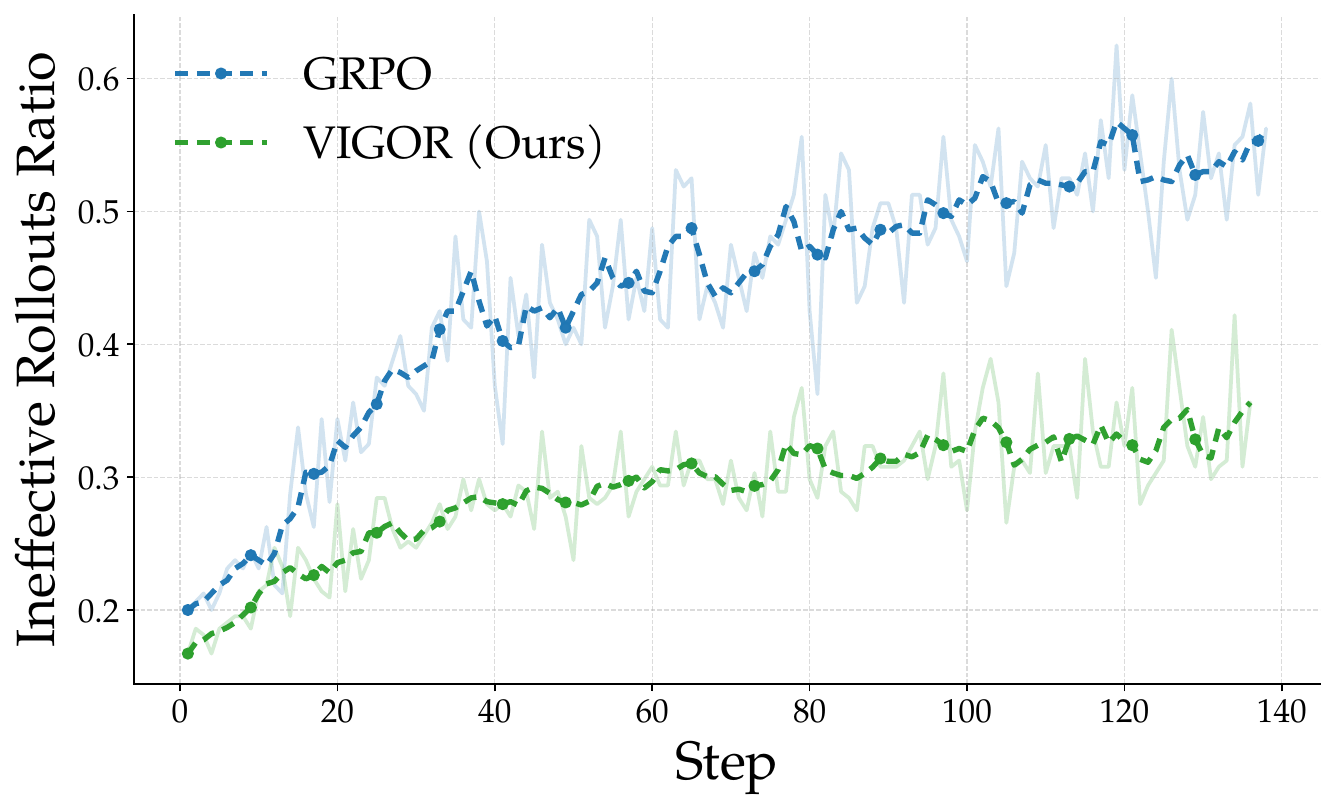}
\end{minipage}
\hspace{0.02\textwidth}
\begin{minipage}[t]{0.4\textwidth}
\centering
\includegraphics[width=\linewidth]{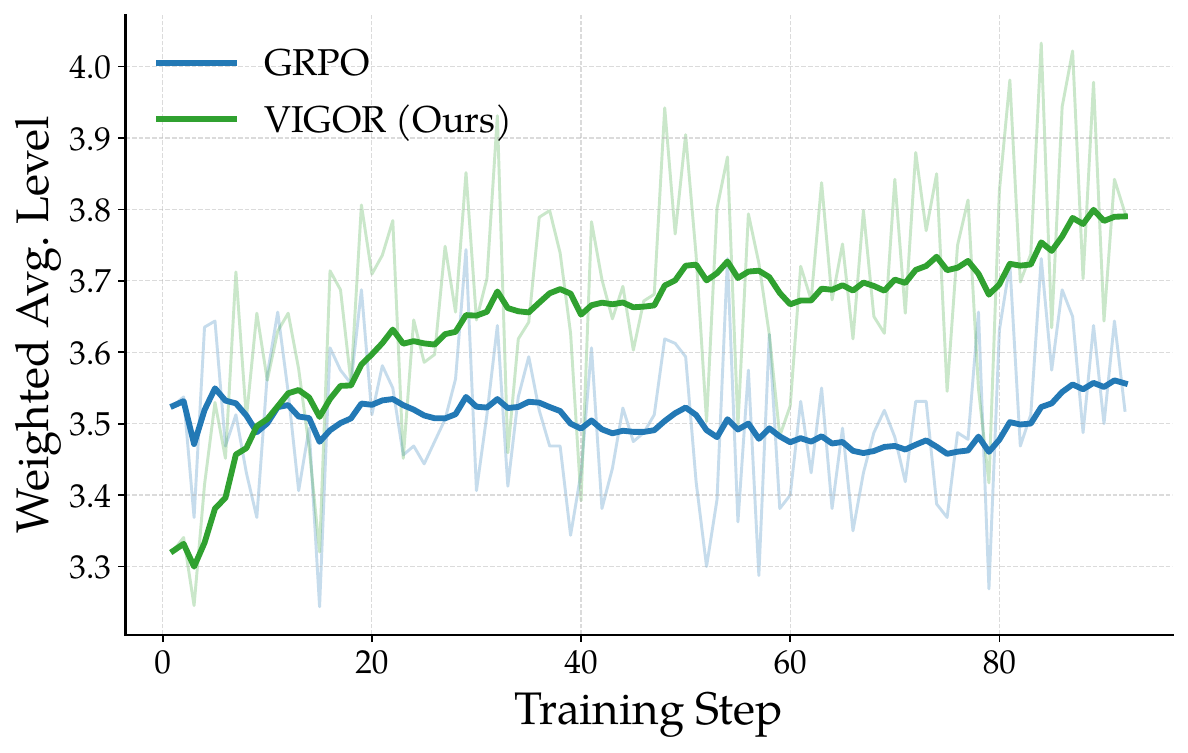}
\end{minipage}
\vspace{-2mm}
\caption{Effective sample ratio and emergent curriculum. \textbf{Left}: Ineffective rollout ratio versus training step. VIGOR substantially reduces the proportion of ineffective samples that contribute zero gradient signal. \textbf{Right}: Rollout-weighted average MATH difficulty over training, showing an emergent curriculum.}
\label{fig:sample_ratio_curriculum}
\end{figure}

\paragraph{VIGOR naturally improves effective sample ratio and rollout efficiency.}
Beyond strengthening the optimization signal through variance, VIGOR's iterative rollout generation naturally filters out ineffective rollouts (groups of rollouts with zero variance), which yield zero advantage and no gradient signal. Unlike dynamic-sampling methods that rely on oversampling, or RL-ZVP, which directly injects additional reward signal, VIGOR uses iterative exploration to progressively allocate more rollouts to prompts that are less likely to produce ineffective rollouts. As shown in the left panel of Figure~\ref{fig:sample_ratio_curriculum}, VIGOR increases the proportion of effective samples that contribute non-trivial learning signal, while reducing rollout waste on already easy or low-information prompts.

\paragraph{VIGOR induces an emergent curriculum.}
Using the Level 1--5 annotations in MATH, we track the rollout-weighted average difficulty,
\begin{equation}
\frac{\sum_i \mathrm{Level}_i \cdot \mathrm{RolloutCount}_i}{\sum_i \mathrm{RolloutCount}_i}.
\end{equation}
As shown in the right panel of Figure~\ref{fig:sample_ratio_curriculum}, the weighted difficulty increases over training, showing that VIGOR naturally reallocates budget toward harder prompts as easier frontier prompts become solved. This curriculum effect does not require explicit difficulty labels or hand-designed scheduling; it emerges from online reward-variance allocation.

\section{Analysis of Selected Prompts}

\noindent
We conduct a case study on training prompts from the MATH dataset using a training run with the Qwen2.5-1.5B model. During training, we track the number of rollouts generated per prompt across multiple passes over the dataset. The aggregate rollout count serves as a proxy for response variance, reflecting how consistently the model produces diverse outcomes for a given prompt.

\begin{wrapfigure}{r}{0.42\textwidth}
\centering
\vspace{-0.8em}
\includegraphics[width=\linewidth]{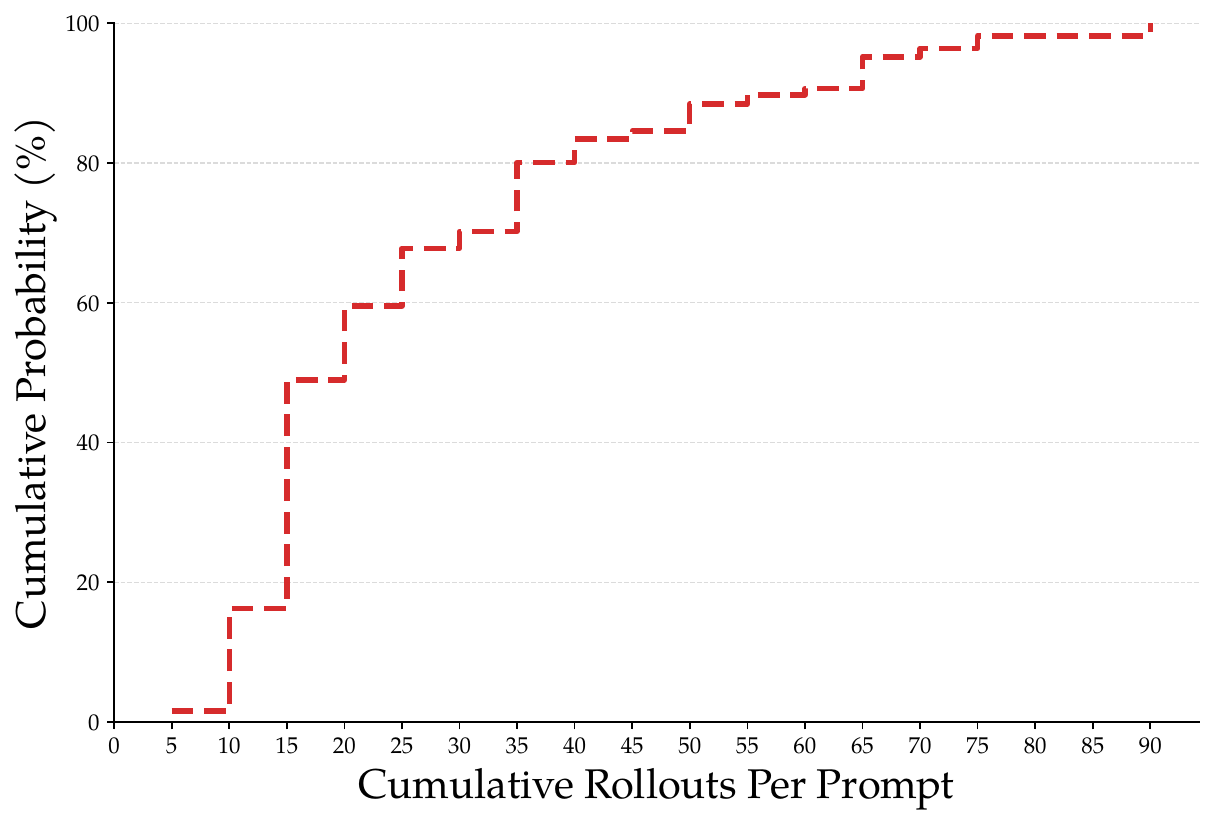}
\caption{Cumulative distribution of rollout counts per prompt.}
\label{fig:appendix-rollout-cdf}
\vspace{-0.8em}
\end{wrapfigure}

Prompts with high selection frequency correspond to the model's learning frontier. These prompts tend to lie at an intermediate difficulty level: they are not trivially solved across all rollouts, yet remain sufficiently interpretable for the model to produce responses that engage with the underlying mathematical structure. In contrast, low-selection prompts are typically either too simple, yielding near-deterministic outputs, or too difficult, resulting in consistently misguided or incorrect responses.

As shown in Figure~\ref{fig:appendix-rollout-cdf}, the rollout-count distribution is strongly right-skewed and its cumulative distribution rises steeply at low rollout counts, indicating that most prompts are not repeatedly selected for additional rollouts across epochs. This suggests that a large fraction of prompts quickly become low-variance, consistent with improved model confidence and reduced need for further exploration.

\begin{tcolorbox}[
colback=blue!5,
colframe=blue!60!black,
title=\textbf{High-Selection Prompts}
]

These prompts were consistently selected for more rollouts, indicating high variance.

\begin{enumerate}\setlength{\leftmargin}{0pt}

\item \textbf{Question:} Triangle $ABC$ has vertices $A(0,8)$, $B(2,0)$, $C(8,0)$. A vertical line intersects $AC$ at $R$ and $\overline{BC}$ at $S$, forming triangle $RSC$. If the area of $\triangle RSC$ is $12.5$, determine the positive difference of the $x$ and $y$ coordinates of point $R$. \\
\textbf{Answer:} 2

\item \textbf{Question:} 4 12-sided dice are rolled. What is the probability that the number of dice showing a two digit number is equal to the number of dice showing a one digit number? Express your answer as a common fraction. \\
\textbf{Answer:} $\frac{27}{128}$

\item \textbf{Question:} Let $a, b, c, d, e, f, g, h$ be real numbers such that $abcd = 4$ and $efgh = 9$. Find the minimum value of $(ae)^2 + (bf)^2 + (cg)^2 + (dh)^2$. \\
\textbf{Answer:} 24

\item \textbf{Question:} What is the largest possible median for the five number set $\{x, 2x, 3, 2, 5\}$ if $x$ can be any integer? \\
\textbf{Answer:} 5

\item \textbf{Question:} On the first day, Barry Sotter used his magic wand to increase an object's length by $\frac{1}{2}$, then by $\frac{1}{3}$, then $\frac{1}{4}$, and so on. On the $n$th day, the length becomes 100 times the original. What is $n$? \\
\textbf{Answer:} 198

\end{enumerate}

\end{tcolorbox}

\vspace{0.5em}

\begin{tcolorbox}[
colback=red!5,
colframe=red!60!black,
title=\textbf{Low-Selection Prompts}
]

These prompts were rarely selected for more rollouts, indicating low variance or low learning signal, either because the model had already mastered the math operation(basic combinatrics, basic calculations), or because the model consistently did not use the correct mathematical principles across all rollouts.

\begin{enumerate}\setlength{\leftmargin}{0pt}

\item \textbf{Question (easy):} Compute $\binom{9}{8}$. \\
\textbf{Answer:} 9

\item \textbf{Question (easy):} How many positive integers $n$ satisfy $(n+8)(n-3)(n-12) < 0$? \\
\textbf{Answer:} 8

\item \textbf{Question (easy):} In the six-digit integer $3A6,792$, what is the largest digit $A$ so that the number is divisible by 3? \\
\textbf{Answer:} 9

\item \textbf{Question (hard):} The rectangle $ABCD$ below has dimensions $AB = 12 \sqrt{3}$ and $BC = 13 \sqrt{3}$. Diagonals $\overline{AC}$ and $\overline{BD}$ intersect at $P$. If triangle $ABP$ is cut out and removed, edges $\overline{AP}$ and $\overline{BP}$ are joined, and the figure is then creased along segments $\overline{CP}$ and $\overline{DP}$, we obtain a triangular pyramid, all four of whose faces are isosceles triangles. Find the volume of this pyramid. \\
\textbf{Answer:} 594

\item \textbf{Question (hard):} An equilateral triangle $PQR$ is inscribed in the ellipse $\frac{x^2}{a^2} + \frac{y^2}{b^2} = 1$, where $Q = (0,b)$ and $\overline{PR}$ is parallel to the $x$-axis. The foci lie on $\overline{QR}$ and $\overline{PQ}$. Find $\frac{PQ}{F_1F_2}$. \\
\textbf{Answer:} $\frac{8}{5}$
\end{enumerate}

\end{tcolorbox}